\documentclass{ecai}  

\usepackage{graphicx}
\usepackage{latexsym}

\usepackage{subcaption}
\usepackage{times}
\usepackage{soul}
\usepackage{url}
\usepackage[hidelinks]{hyperref}
\usepackage[utf8]{inputenc}
\usepackage{booktabs}
\usepackage{algorithm}
\usepackage{algorithmic}
\usepackage{multirow,array}
\usepackage{amstext,amsmath,amssymb,amsthm}
\usepackage{bm}
\usepackage{xcolor}
\usepackage{nicefrac}
\newtheorem{proposition}{Proposition}
\newtheorem{Theorem}{Theorem}
\allowdisplaybreaks

\begin{document}

\begin{frontmatter}

\title{Offline Decentralized Multi-Agent \\Reinforcement Learning}


\author[A]{\fnms{Jiechuan}~\snm{Jiang}}
\author[A]{\fnms{Zongqing}~\snm{Lu}\thanks{Corresponding Author. Email: zongqing.lu@pku.edu.cn}}

\address[A]{Peking University}

\begin{abstract}
In many real-world multi-agent cooperative tasks, due to high cost and risk, agents cannot continuously interact with the environment and collect experiences during learning, but have to learn from offline datasets. However, the transition dynamics in the dataset of each agent can be much different from the ones induced by the learned policies of other agents in execution, creating large errors in value estimates. Consequently, agents learn uncoordinated low-performing policies. In this paper, we propose a framework for offline decentralized multi-agent reinforcement learning, which exploits \textit{value deviation} and \textit{transition normalization} to deliberately modify the transition probabilities. Value deviation optimistically increases the transition probabilities of high-value next states, and transition normalization normalizes the transition probabilities of next states. They together enable agents to learn high-performing and coordinated policies. Theoretically, we prove the convergence of Q-learning under the altered \textit{non-stationary} transition dynamics. Empirically, we show that the framework can be easily built on many existing offline reinforcement learning algorithms and achieve substantial improvement in a variety of multi-agent tasks.
\end{abstract}

\end{frontmatter}

\section{Introduction}

Reinforcement learning (RL) has shown great potential in domains, including recommendation systems \cite{li2010contextual}, games \cite{vinyals2019grandmaster}, and robotics \cite{ibarz2021how}. When applying RL to real-world applications, the agent needs to continuously interact with the environment and collect the experiences for learning, which however can be costly, risky, and time-consuming. One way to address this is offline RL, where the agent learns the policy, without interacting with the environment, from a fixed dataset of experiences collected by any behavior policy. The main challenge of offline RL is the \textit{extrapolation error}, an error incurred by the mismatch between the experience distributions of the learned policy and dataset \cite{fujimoto2019off}. Many offline RL methods \cite{fujimoto2019off,kumar2019stabilizing,wu2019behavior,kumar2020conservative} have been proposed to address the value overestimate caused by the mismatch between action distributions of the learned policy and behavior policy. However, much less attention has been paid to the mismatch between the transition dynamics in the dataset and the real ones in execution. The main reason is that the mismatch of transition dynamics is negligible given a large dataset, if the environment is stationary. Nevertheless, in many real-world scenarios, there are \textit{other} learning agents in the environment, \textit{e.g.}, autonomous driving. That means the policies of other agents during data collection can be significantly different from their policies in execution, which creates the mismatch of transition dynamics and hence undermines the learned policy.


More specifically, these scenarios can be formulated as the offline decentralized multi-agent setting, where agents cooperate on the task but each learns the policy from its own fixed dataset. The dataset of each agent is independently collected by any behavior policies \textit{arbitrarily} (\textit{i.e.}, we do not make any assumptions on data collection) and contains its own action instead of the joint action of all agents. This setting resembles many industrial applications where agents belong to different companies and the actions of other agents are not accessible, \textit{e.g.}, autonomous vehicles or robots. Apparently, this setting does not fit the paradigm of offline centralized training and decentralized execution (CTDE) in multi-agent reinforcement learning (MARL)\footnote{Even if the datasets of all agents can be accessed in a centralized way, a full dataset that contains the joint actions cannot be constructed, because the datasets are fully independent without any common information.} \cite{yang2021believe}, 
and each agent has to learn a cooperative policy in an \emph{offline and fully decentralized} way. In such decentralized multi-agent settings, from the perspective of an individual agent, other agents are a part of the environment, thus the transition dynamics experienced by each agent depend on the policies of other agents and change as other agents update their policies \cite{foerster2017stabilising}. As the learned policies of other agents will be inconsistent with their behavior policies, the transition dynamics induced by the learned policies of other agents in execution will be different from the transition dynamics in the dataset, causing a large mismatch between the transition dynamics. This mismatch can lead to a low-performing policy for each agent. Furthermore, as agents learn in a decentralized way on the datasets with different distributions of experiences collected by various behavior policies, the estimated values of the same state can be much different between agents, which causes the learned policies cannot well coordinate with each other.

In this paper, we propose a framework for offline fully decentralized multi-agent reinforcement learning, where, to overcome the mismatch between transition dynamics and miscoordination, \textit{value deviation} and \textit{transition normalization} are introduced to deliberately modify the transition probabilities in the dataset. 
During data collection, if one agent takes a `good' action while other agents take `bad' actions at a state, the transition probabilities of low-value next states will be high. 
Thus, the Q-value of the `good' action will be \textit{underestimated} by the agent. 
As other agents are also learning, their learned policies are expected to become better than their behavior policies. 
Therefore, for each agent, the transition probabilities of high-value next states in execution will be higher than those in the dataset. 
So, we let each agent be \emph{optimistic} towards other agents and multiply the transition probabilities by the deviation of the value of next state from the expected value over all next states, to make the estimated transition probabilities \textit{close} to the transition probabilities induced by the learned policies of other agents. 
To address the miscoordination caused by diverse value estimates of agents, we normalize the transition probabilities in the dataset to be uniform. Transition normalization helps to build the consensus about value estimate among agents and hence promotes coordination. 
By combining value deviation and transition normalization, agents are able to learn high-performing and coordinated policies in an offline and fully decentralized way.

Value deviation and transition normalization make the transition dynamics non-stationary during learning. However, we mathematically prove the convergence of Q-learning under such purposely controlled non-stationary transition dynamics. Moreover, by derivation, we show that value deviation and transition normalization can take effect \textit{merely} as the weights of the objective function, thus our framework can be easily built on many existing offline single-agent RL algorithms that address the overestimation incurred by out-of-distribution actions. Empirically, we provide an example instantiation and the thorough analysis of our framework on BCQ \cite{fujimoto2019off}, termed MABCQ, and also test the variants on CQL \cite{kumar2020conservative} and TD3+BC \cite{fujimoto2021minimalist}, termed MACQL and MATD3+BC respectively. Experimental results show that our method substantially outperforms baselines in a variety of multi-agent tasks, including multi-agent mujoco \cite{de2020deep}, SMAC \cite{samvelyan19smac} and MPE \cite{lowe2017multi}. To the best of our knowledge, this is the \textit{first} work for offline and decentralized multi-agent reinforcement learning.


\section{Preliminaries}
\label{sec:preliminaries}

We consider $N$ agents in multi-agent MDP \cite{oliehoek2016concise} $M_{\mathrm{env}}= < \mathcal{S}, \mathcal{A}, R, P_{\mathrm{env}}, \gamma>$ with the state space $\mathcal{S}$ and the joint action space $\mathcal{A}$. At each timestep, each agent $i$ gets state\footnote{Global state is only for the convenience of theoretical analysis \cite{kuba2022trust} the agents could learn based on partial observation in practice.} $s$, and performs an individual action $a_i$, and the environment transitions to the next state $s'$ by taking the joint action $\boldsymbol{a}$ with the transition probability $P_{\mathrm{env}}\left(s^{\prime} | s, \boldsymbol{a}\right)$. All agents get a shared reward $r = R\left(s\right)$, which is simplified to just depending on state \cite{schulman2015trust}\footnote{This simplification is only for the convenience of theoretical analysis, but not necessary in practice.}. All the agents learn to maximize the expected return $\mathbb{E} \sum_{t=0}^{\infty} \gamma^{t} r_{t}$, where $\gamma$ is a discount factor. However, in the fully decentralized learning, $M_{\mathrm{env}}$ is partially observable to each agent since each agent only observes its own action $a_i$ instead of the joint action $\boldsymbol{a}$. During execution, from the perspective of each agent $i$, there is an experienced MDP $M_{\mathcal{E}_i} = <\mathcal{S}, \mathcal{A}_i, R, P_{\mathcal{E}_i}, \gamma>$ with the individual action space $\mathcal{A}_i$ and the \textbf{online transition probability} 
\begin{equation}
\notag
\setlength{\abovedisplayskip}{3pt}
P_{\mathcal{E}_i}\left(s^{\prime} | s, a_i\right)={\sum}_{\boldsymbol{a}_{-i}}P_{\mathrm{env}}\left(s^{\prime} | s, a_i,\boldsymbol{a}_{-i}\right)  \boldsymbol{\pi}_{\mathcal{E}_{-i}}(\boldsymbol{a}_{-i} | s),
\setlength{\belowdisplayskip}{2pt}
\end{equation}
where $a_{-i}$ and $\boldsymbol{\pi}_{\mathcal{E}_{-i}}$ respectively denote the joint action and the learned joint policy of all agents except agent $i$. However, if the agent cannot interact with other agents in the environment, $P_{\mathcal{E}_i}$ is \textit{unknown}. 

In \textit{offline decentralized settings}, each agent $i$ only has access to a fixed offline dataset $\mathcal {B}_i$, which is pre-collected by behavior policies $\boldsymbol{\pi}_{\mathcal{B}}$ and contains the tuples $(s,a_i,r,s')$. As defined in BCQ \cite{fujimoto2019off}, the visible MDP $M_{\mathcal {B}_i}= <\mathcal{S}, \mathcal{A}_i, R, P_{\mathcal {B}_i}, \gamma>$ is constructed on $\mathcal {B}_i$, which has the \textbf{offline transition probability}

\begin{equation}
\notag
\setlength{\abovedisplayskip}{3pt}
P_{\mathcal{B}_i}\left(s^{\prime} | s, a_i\right)={\sum}_{\boldsymbol{a}_{-i}}P_{\mathrm{env}}\left(s^{\prime} | s, a_i,\boldsymbol{a}_{-i}\right)  \boldsymbol{\pi}_{\mathcal{B}_{-i}}(\boldsymbol{a}_{-i} | s),
\end{equation}
As the learned policies of other agents $\boldsymbol{\pi}_{\mathcal{E}_{-i}}$ may greatly deviate from their behavior policies $\boldsymbol{\pi}_{\mathcal{B}_{-i}}$, $P_{\mathcal {B}_i}$ can be largely biased from $P_{\mathcal{E}_i}$. We define the discrepancy between $P_{\mathcal {B}_i}$ and $P_{\mathcal{E}_i}$ as \textbf{\textit{transition bias}}. The large extrapolation errors caused by transition bias, and the differences in value estimates between agents lead to uncoordinated low-performing policies.
\begin{table}
	\setlength{\extrarowheight}{3pt}
	\centering
	\setlength{\abovecaptionskip}{-0.1cm}
	\caption{The matrix game.}
	\label{tab:1step_game}
        \vspace{1mm}
	\begin{small}
		\begin{tabular}{llcc}
			&                                                       & \multicolumn{2}{c}{{\color[HTML]{00009B} Agent $2$}}                                                        \\
			&                                                       & \multicolumn{1}{l}{{\color[HTML]{00009B} $a_1$ ($0.4$)}} & \multicolumn{1}{l}{{\color[HTML]{00009B} $a_2$ ($0.6$)}} \\ \cline{3-4} 
			\multicolumn{1}{c}{{\color[HTML]{9A0000} }}                            & \multicolumn{1}{c|}{{\color[HTML]{9A0000} $a_1$ ($0.8$)}} & \multicolumn{1}{c|}{1}                               & \multicolumn{1}{c|}{5}                               \\ \cline{3-4} 
			\multicolumn{1}{c}{\multirow{-2}{*}{{\color[HTML]{9A0000} Agent $1$}}} & \multicolumn{1}{c|}{{\color[HTML]{9A0000} $a_2$ ($0.2$)}} & \multicolumn{1}{c|}{6}                               & \multicolumn{1}{c|}{1}                               \\ \cline{3-4} 
		\end{tabular}
	\end{small}
\end{table}

\begin{table}
	\setlength{\extrarowheight}{3pt}
	\centering
	\setlength{\abovecaptionskip}{-0.15cm}
	\setlength{\belowcaptionskip}{-0.25cm}
	\caption{Transition probabilities and expected returns calculated in the dataset.}
	\label{tab:step_game_return}
        \begin{small}
		\begin{tabular}{cccc}
			\multicolumn{1}{l}{} & action & transition  & expected return        \\ \cline{2-4} 
			\multicolumn{1}{c|}{{\color[HTML]{9A0000} }}                            & \multicolumn{1}{c|}{{\color[HTML]{9A0000} }}                        & \multicolumn{1}{c|}{$p(1 | {\color[HTML]{9A0000} a_1}) = 0.4$} & \multicolumn{1}{c|}{}                      \\ \cline{3-3}
			\multicolumn{1}{c|}{{\color[HTML]{9A0000} }}                            & \multicolumn{1}{c|}{\multirow{-2}{*}{{\color[HTML]{9A0000} $a_1$}}} & \multicolumn{1}{c|}{$p(5 | {\color[HTML]{9A0000} a_1}) = 0.6$} & \multicolumn{1}{c|}{\multirow{-2}{*}{3.4}} \\ \cline{2-4} 
			\multicolumn{1}{c|}{{\color[HTML]{9A0000} }}                            & \multicolumn{1}{c|}{{\color[HTML]{9A0000} }}                        & \multicolumn{1}{c|}{$p(6 | {\color[HTML]{9A0000} a_2}) = 0.4$} & \multicolumn{1}{c|}{}                      \\ \cline{3-3}
			\multicolumn{1}{c|}{\multirow{-4}{*}{\rotatebox{90}{\color[HTML]{9A0000} Agent $1$}}} & \multicolumn{1}{c|}{\multirow{-2}{*}{{\color[HTML]{9A0000} $a_2$}}} & \multicolumn{1}{c|}{$p(1 | {\color[HTML]{9A0000} a_2}) = 0.6$} & \multicolumn{1}{c|}{\multirow{-2}{*}{3.0}} \\ \cline{2-4} 
		\end{tabular}
		\begin{tabular}{cccc}
			\multicolumn{1}{l}{}                                                    & action                                                              & transition                                 & expected return                            \\ \cline{2-4} 
			\multicolumn{1}{c|}{{\color[HTML]{00009B} }}                            & \multicolumn{1}{c|}{{\color[HTML]{00009B} }}                        & \multicolumn{1}{c|}{$p(1 | {\color[HTML]{00009B} a_1}) = 0.8$} & \multicolumn{1}{c|}{}                      \\ \cline{3-3}
			\multicolumn{1}{c|}{{\color[HTML]{00009B} }}                            & \multicolumn{1}{c|}{\multirow{-2}{*}{{\color[HTML]{00009B} $a_1$}}} & \multicolumn{1}{c|}{$p(6 | {\color[HTML]{00009B} a_1}) = 0.2$} & \multicolumn{1}{c|}{\multirow{-2}{*}{2.0}} \\ \cline{2-4} 
			\multicolumn{1}{c|}{{\color[HTML]{00009B} }}                            & \multicolumn{1}{c|}{{\color[HTML]{00009B} }}                        & \multicolumn{1}{c|}{$p(5 | {\color[HTML]{00009B} a_2}) = 0.8$} & \multicolumn{1}{c|}{}                      \\ \cline{3-3}
			\multicolumn{1}{c|}{\multirow{-4}{*}{\rotatebox{90}{\color[HTML]{00009B} Agent $2$}}} & \multicolumn{1}{c|}{\multirow{-2}{*}{{\color[HTML]{00009B} $a_2$}}} & \multicolumn{1}{c|}{$p(1 | {\color[HTML]{00009B} a_2}) = 0.2$} & \multicolumn{1}{c|}{\multirow{-2}{*}{4.2}} \\ \cline{2-4} 
		\end{tabular}
	\end{small}
\end{table}

To intuitively illustrate the issue, we introduce a two-player matrix game as depicted in Table~\ref{tab:1step_game}, where the action distributions of the behavior policies of the two agents are $[0.8,0.2]$ and $[0.4,0.6]$, respectively. 
Table~\ref{tab:step_game_return} shows the transition probabilities and expected returns calculated independently by the agents from the datasets collected by the behavior policies. 
However, as the behavior policies are poor, when agent 1 chooses the optimal action {\color[HTML]{9A0000} $a_2$}, agent 2 has a higher probability to select the suboptimal action {\color[HTML]{00009B} $a_2$}, which leads to low transition probabilities of \textit{true} high-value next states ($p(6|{\color[HTML]{9A0000} a_2})=0.4$ vs. $p(1|{\color[HTML]{9A0000} a_2})=0.6$). 
So the agents underestimate the optimal actions and converge to the suboptimal policies $({\color[HTML]{9A0000} a_1}, {\color[HTML]{00009B} a_2})$, rather than the optimal policy $({\color[HTML]{9A0000} a_2}, {\color[HTML]{00009B} a_1})$.

\section{Offline Decentralized MARL Framework}
\label{sec:method}

In fully decentralized learning, as the policies of other agents are not accessible, it is hard for an agent to learn a policy that can coordinate well with other agents, merely from the dataset that only contains its own action. To tackle this challenging problem, we propose a framework, which newly introduces value deviation and transition normalization to address transition bias and miscoordination, and leverages the offline single-agent RL algorithm to avoid out-of-distribution actions. The convergence under the non-stationary transition dynamics is theoretically guaranteed, and an example instantiation of the framework is provided.

\subsection{Value Deviation}

If the behavior policies of some agents are low-performing during data collection, they usually take `bad' actions to cooperate with the `good' actions of other agents, which leads to high transition probabilities of low-value next states. When agent $i$ performs Q-learning with the dataset $\mathcal {B}_i$, the Bellman operator $\mathcal{T}$ is approximated by the transition probability $P_{\mathcal {B}_i}\left(s^{\prime} | s, a_i\right)$ to estimate the expectation over $s'$:
\begin{equation}
\notag
\setlength{\abovedisplayskip}{3pt}
\mathcal{T} Q_i(s, a_i)=\mathbb{E}_{s^{\prime} \sim P_{\mathcal {B}_i}\left(s^{\prime} | s, a_i\right)}\left[r+\gamma \max _{\hat{a}_i} Q_i\left(s^{\prime}, \hat{a}_i\right)\right].
\setlength{\belowdisplayskip}{3pt}
\end{equation}
If $P_{\mathcal {B}_i}$ of a high-value $s'$ is lower than $P_{\mathcal{E}_i}$, the Q-value of the state-action pair $(s,a_i)$ is underestimated, which will cause large extrapolation error. 

However, as the policies of other agents are also updating towards maximizing the Q-values, $P_{\mathcal{E}_i}$ of high-value next states will grow higher than $P_{\mathcal {B}_i}$. Thus, we let each agent be optimistic towards other agents and modify $P_{\mathcal {B}_i}$ as
\begin{equation}
\notag
\setlength{\abovedisplayskip}{3pt}
P_{\mathcal {B}_i}\left(s^{\prime} | s, a_i\right)\cdot(\underset{\text{value deviation}}{\underbrace{1+\frac{V_i^*(s')-\mathbb{E}_{\hat{s}'}V_i^*(\hat{s}')}{|\mathbb{E}_{\hat{s}'}V_i^*(\hat{s}')|}}})\cdot\frac{1}{z_i^{vd}},
\setlength{\belowdisplayskip}{3pt}
\end{equation}
where the state value $V_i^{*}(s)=\max _{a_i} Q_i(s, a_i)$, $1+\nicefrac{V_i^*(s')-\mathbb{E}_{\hat{s}'}V_i^*(\hat{s}')}{|\mathbb{E}_{\hat{s}'}V_i^*(\hat{s}')|}$ is the deviation of the value of next state from the expected value over all next states, which increases the transition probabilities of high-value next states and decreases those of low-value next states, and $z_i^{vd} = \sum_{s'}P_{\mathcal {B}_i}\left(s^{\prime} | s, a_i\right)*(1+\nicefrac{V_i^*(s')-\mathbb{E}_{\hat{s}'}V_i^*(\hat{s}')}{|\mathbb{E}_{\hat{s}'}V_i^*(\hat{s}')|})$ is a normalization term to make sure the sum of the transition probabilities is one. \textit{Value deviation} modifies the transition probabilities to be close to $P_{\mathcal{E}_i}$ and hence reduces the transition bias. The optimism towards other agents helps the agent discover potential good actions which are hidden by the poor behavior policies of other agents.

\subsection{Transition Normalization}


As $\mathcal {B}_i$ of each agent is individually collected by different behavior policies, the diverse combinations of behavior policies of all agents lead to the value of the same state $s$ being overestimated by some agents, while being underestimated by others. Since the agents are trained to reach high-value states, the large disagreement on state values will cause miscoordination of the learned policies. To overcome the problem, we normalize $P_{\mathcal {B}_i}$ to be uniform over next states,
\begin{equation}
\notag
\setlength{\abovedisplayskip}{3pt}
P_{\mathcal {B}_i}\left(s^{\prime} | s, a_i\right)\cdot\underset{\text{transition normalization}}{\underbrace{\frac{1}{P_{\mathcal {B}_i}\left(s^{\prime} | s, a_i\right)}}}\cdot\frac{1}{z_i^{tn}}\footnote{Although the modification can be simply written as $\nicefrac{1}{z_i^{tn}}$, we present it in such a way that we can later conveniently combine transition normalization with value deviation.}
\setlength{\belowdisplayskip}{3pt}
\end{equation}
where $z_i^{tn}$ is a normalization term that is the number of different $s'$ given $(s,a_i)$ in $\mathcal{B}_i$. \textit{Transition normalization} enforces that each agent has the same $P_{\mathcal {B}_i}$ when it acts the learned action $a_i^*$ on the same state $s$, and we have the following proposition.

\begin{proposition}
	In episodic environments, if each agent $i$ performs Q-learning on $\mathcal {B}_i$, all agents will converge to the same $V^*$ if they have the same transition probability on any state where each agent $i$ acts the learned action $a_i^*$.
\end{proposition}

The proof is provided in Appendix\footnote{Appendix is available at \url{https://arxiv.org/abs/2108.01832}.}. This proposition implies that transition normalization can enable agents to have the same state value estimate. However, to satisfy $P_{\mathcal {B}_{1}}\left(s^{\prime} | s, a_{1}^{*}\right) = P_{\mathcal {B}_{2}}\left(s^{\prime} | s, a_{2}^{*}\right) = \ldots = P_{\mathcal {B}_{N}}\left(s^{\prime} | s, a_{N}^{*}\right)$ for all $s' \in \mathcal{S}$ in the datasets, the agents should have the same set of $s'$ at $(s,a^*)$, which is a strong assumption. In practice, although the assumption is not strictly satisfied, transition normalization can still normalize the transition probabilities, encouraging the estimated state value $V^*$ to be close to each other.

\subsection{Optimization Objective}

Combining value deviation $1+\nicefrac{(V_i^*(s')-\mathbb{E}_{\hat{s}'}V_i^*(\hat{s}'))}{|\mathbb{E}_{\hat{s}'}V_i^*(\hat{s}')|}$, denoted as $\lambda_{{vd}_i}$, and transition normalization $\nicefrac{1}{P_{\mathcal {B}_i}\left(s^{\prime} | s, a_i\right)}$, denoted as $\lambda_{{tn}_i}$, we modify $P_{\mathcal {B}_i}$ as
\begin{equation}
\notag
\setlength{\abovedisplayskip}{3pt}
\hat{P}_{\mathcal {B}_i}\left(s^{\prime} | s, a_i\right) = P_{\mathcal {B}_i}\left(s^{\prime} | s, a_i\right)\cdot\frac{\lambda_{{tn}_i}\lambda_{{vd}_i}}{z_i},
\setlength{\belowdisplayskip}{3pt}
\end{equation}
where $z_i = \sum_{s'}(1+\nicefrac{(V_i^*(s')-\mathbb{E}_{\hat{s}'}V_i^*(\hat{s}'))}{|\mathbb{E}_{\hat{s}'}V_i^*(\hat{s}')|})$ is the normalization term. 
Indeed, $\hat{P}_{\mathcal {B}_i}$ makes offline learning on $\mathcal{B}_i$ similar to \textit{online} decentralized MARL. 
In the initial stage, $\lambda_{{vd}_i}$ is close to $1$ since $Q_i(s,a_i)$ is not learned yet, and the transition probabilities are uniform, resembling the start stage in online learning where all agents are acting randomly. 
During training, the transition probabilities of high-value states gradually grow by value deviation, which is an analogy of other agents improving their policies in online learning. 
Therefore, $\hat{P}_{\mathcal {B}_i}$ encourages the agents to learn high-performing policies and improves coordination. 

Although $\hat{P}_{\mathcal {B}_i}$ is non-stationary (\textit{i.e.}, $\lambda_{{vd}_i}$ changes along with the updates of Q-value), we have the following theorem that guarantees the convergence of Bellman operator $\mathcal{T}$ under $\hat{P}_{\mathcal {B}_i}$,
\[\mathcal{T} Q_i(s, a_i)=\mathbb{E}_{s' \sim \hat{P}_{\mathcal {B}_i}\left(s^{\prime} | s, a_i\right)}\left[r+\gamma \max _{\hat{a}_i} Q_i\left(s^{\prime}, \hat{a}_i\right)\right].\]

\begin{Theorem}\label{theorem1}
	Under the non-stationary transition probability $\hat{P}_{\mathcal {B}_i}$, the Bellman operator $\mathcal{T}$ is a contraction and converges to a unique fixed point when $\gamma < \nicefrac{r_{\min}}{2r_{\max} - r_{\min}}$, if the reward is bounded by the positive region $[r_{\min},r_{\max}]$.
\end{Theorem}



The proof is provided in Appendix. As any positive affine transformation of the reward function does not change the optimal policy in the fixed-horizon environments \cite{zhang2021brac}, Theorem \ref{theorem1} holds in general, and we can rescale the reward to make $r_{\min}$ arbitrarily close to $r_{\max}$ so as to make the upper bound of $\gamma$ \textit{close} to $1$.

In deep reinforcement learning, directly modifying the transition probability is infeasible. However, we can instead modify the sampling probability to achieve the same effect. The optimization objective of decentralized deep Q-learning
$\mathbb{E}_{p_{\mathcal{B}_i}(s,a_i,s')}|Q_i(s,a_i)-y_i|^2$
is calculated by sampling the batch from $\mathcal{B}_i$ according to the sampling probability $p_{\mathcal{B}_i}(s,a_i,s')$. By factorizing $p_{\mathcal{B}_i}(s,a_i,s')$, we have
\[ \underset{ {\color[HTML]{9A0000}\text{sampling probability}} }{\underbrace{p_{\mathcal{B}_i}(s,a_i,s')}}  = p_{\mathcal{B}_i}(s,a_i) \cdot\underset{ {\color[HTML]{00009B}\text{transition probability} }}{\underbrace{P_{\mathcal{B}_i}(s' | s,a_i) }}.\]
Therefore, we can modify the transition  probability as $\frac{\lambda_{{tn}_i}\lambda_{{vd}_i}}{z_i}P_{\mathcal {B}_i}\left(s^{\prime} | s, a_i\right)$ and scale $p_{\mathcal{B}_i}(s,a_i)$ with $z_i$. Then, the sampling probability can be re-written as
\[ \underset{ {\color[HTML]{9A0000}\text{modified sampling probability}} }{\underbrace{\lambda_{{tn}_i}\lambda_{{vd}_i}p_{\mathcal{B}_i}(s,a_i,s')}}  = z_i p_{\mathcal{B}_i}(s,a_i) \cdot\underset{{\color[HTML]{00009B}\text{modified transition probability} }}{\underbrace{\frac{\lambda_{{tn}_i}\lambda_{{vd}_i}}{z_i}P_{\mathcal{B}_i}(s' | s,a_i)}}.\]
Since $z_i$ is independent of $s'$, it can be regarded as a scale factor on $p_{\mathcal{B}_i}(s,a_i)$, which will not change the expected target value $y_i$. Thus, sampling batches according to the modified sampling probability can achieve the same effect as modifying the transition probability. Using importance sampling, the modified optimization objective is
\[\begin{split}
&\mathbb{E}_{ \lambda_{{tn}_i}\lambda_{{vd}_i}p_{\mathcal{B}_i}(s,a_i,s')}|Q_i(s,a_i)-y_i|^2\\
&=\mathbb{E}_{p_{\mathcal{B}_i}(s,a_i,s')}\frac{\lambda_{{tn}_i}\lambda_{{vd}_i}p_{\mathcal{B}_i}(s,a_i,s')}{p_{\mathcal{B}_i}(s,a_i,s')}|Q_i(s,a_i)-y_i|^2\\
&=\mathbb{E}_{p_{\mathcal{B}_i}(s,a_i,s')}\lambda_{{tn}_i}\lambda_{{vd}_i}|Q_i(s,a_i)-y_i|^2.
\end{split}\]
We can see that $\lambda_{{tn}_i}$ and $\lambda_{{vd}_i}$ simply take effect as the weights of the objective function, which makes them easily integrated with existing offline RL methods.

\begin{algorithm}[!t]
	\caption{MABCQ}
	\label{alg:1}
	\begin{algorithmic}[1]
		\FOR{$i \in N$}	
		
		\STATE Initialize the conditional VAEs:			
		
		$G^{1}_i=\{E^{1}_i\left(\mu^{1}, \sigma^{1} | s, a\right), D^{1}_i\left(a | s, z^{1} \right)\}$, \\			
		$G^{2}_i=\{E^{2}_i\left(\mu^{2}, \sigma^{2} | s, a, s'\right), D^{2}_i\left(a | s, s', z^{2} \right)\}$.		
		
		\STATE Initialize Q-network $Q_i$, perturbation network $\xi_i$, and their target networks $\hat{Q}_i$ and $\hat{\xi}_i$.				
		\STATE Fit the VAEs $G^{1}_i$ and $G^{2}_i$ using $\mathcal{B}_i$.			
		\FOR{$t = 1, \ldots, max\_update$}		
		\STATE Sample a mini-batch from $\mathcal{B}_i$.
		\STATE Update $Q_i$ by minimizing (\ref{eq:1}).
		\STATE Update $\xi_i$ by maximizing (\ref{eq:2}).
		\STATE Update the target networks $\hat{Q}_i$ and $\hat{\xi}_i$.		
		\ENDFOR		
		\ENDFOR	
	\end{algorithmic}
\end{algorithm}
\subsection{An Example Instantiation}

Our framework can be practically instantiated on many offline single-agent RL algorithms that address the overestimation incurred by out-of-distribution actions. Here, we give the instantiation of the framework on BCQ \cite{fujimoto2019off}, termed MABCQ. To make MABCQ adapt to high-dimensional continuous spaces, for each agent $i$, we train a Q-network $Q_i$, a perturbation network $\xi_i$, and a conditional VAE $G^{1}_i=\{E^{1}_i\left(\mu^{1}, \sigma^{1} | s, a\right), D^{1}_i\left(a | s, z^{1} \sim\left(\mu^{1}, \sigma^{1}\right)\right)\}$. In execution, each agent $i$ generates $n$ actions by $G_i^1$, adds small perturbations $\in [-\Phi ,\Phi ]$ on the actions using $\xi_i$, and then selects the action with the highest value in $Q_i$. The policy can be written as
\[\begin{split}
\pi_i(s)=\underset{a_i^{j}+\xi_i\left(s, a_i^{j}\right)}{\operatorname{argmax}} Q_i\left(s, a_i^{j}+\xi\left(s, a_i^{j}\right)\right), \\
\text{where } \left\{a_i^{j} \sim G_i^1(s)\right\}_{j=1}^{n}.
\end{split}\]
$Q_i$ is updated by minimizing 
\begin{equation}
\label{eq:1}
\begin{split}
\mathbb{E}_{p_{\mathcal{B}_i}(s,a_i,s')}\lambda_{{tn}_i}\lambda_{{vd}_i}|Q_i(s,a_i)-y_i|^2, \\
\text{where } y_i=r + \gamma \hat{Q}_i(s',\hat{\pi}_i(s')).
\end{split}
\end{equation}
$y_i$ is calculated by the target networks $\hat{Q}_i$ and $\hat{\xi}_i$, where $\hat{\pi}_i$ is correspondingly the policy induced by $\hat{Q}_i$ and $\hat{\xi}_i$. 
$\xi_i$ is updated by maximizing 
\begin{equation}
\setlength{\abovedisplayskip}{5pt}
\label{eq:2}
\mathbb{E}_{p_{\mathcal{B}_i}(s,a_i,s')}\lambda_{{tn}_i}\lambda_{{vd}_i}Q_i\left(s, a_i+\xi_i\left(s, a_i\right)\right).
\setlength{\belowdisplayskip}{5pt}
\end{equation}

To estimate $\lambda_{{vd}_i}$, we need $V_i^*(s') =  \hat{Q}_i(s',\hat{\pi}_i(s'))$ and $\mathbb{E}_{s^{\prime}} [V_i^{*}\left(s^{\prime}\right)]=\frac{1}{\gamma}(\hat{Q}_i(s, a_i)-r)$, which can be estimated from the sample without actually going through all $s^{\prime}$. We estimate $\lambda_{{vd}_i}$ using the target networks to stabilize $\lambda_{{vd}_i}$ along with the updates of $Q_i$ and $\xi_i$. To avoid extreme values, we clip $\lambda_{{vd}_i}$ to the region $[1-\epsilon, 1+\epsilon]$, where $\epsilon$ is the optimism level.

To estimate $\lambda_{{tn}_i}$, we train a VAE $G^{2}_i=\{E^{2}_i\left(\mu^{2}, \sigma^{2} | s, a, s'\right), D^{2}_i\left(a | s, s', z^{2} \sim\left(\mu^{2}, \sigma^{2}\right)\right)\}$. Since the latent variable of VAE follows the Gaussian distribution, we use the mean as the encoding of the input and estimate the probability density functions: $\rho_i (s,a) \approx  \rho_{\mathcal{N}(0,1)}(\mu^1_i)$ and $\rho_i (s,a,s') \approx  \rho_{\mathcal{N}(0,1)}(\mu^2_i)$, where $\rho_{\mathcal{N}(0,1)}$ is the density of unit Gaussian distribution. The conditional density is $\rho_i (s' | a,s) \approx \nicefrac{\rho_{\mathcal{N}(0,1)}(\mu^2_i)}{\rho_{\mathcal{N}(0,1)}(\mu^1_i)}$ and the transition probability is $P_{\mathcal{B}_i}(s' | s,a_i) \approx \int_{s' - \frac{1}{2}\delta_{\mathcal{S}} }^{s' + \frac{1}{2}\delta_{\mathcal{S}} } \rho_i(s' | s, a) \mathrm{d} s' \approx \rho_i(s' | s, a)\left \| \delta_{\mathcal{S}}\right \|$ when  $\left \| \delta_{\mathcal{S}}\right \|$ is a small constant. Approximately,  we have 
\begin{equation}
\notag
\setlength{\abovedisplayskip}{2pt}
\lambda_{{tn}_i} = \frac{\rho_{\mathcal{N}(0,1)}(\mu^1_i)}{\rho_{\mathcal{N}(0,1)}(\mu^2_i)},
\setlength{\belowdisplayskip}{3pt}
\end{equation}
and the constant $\left \| \delta_{\mathcal{S}}\right \|$ is considered in $z_i$. In practice, we find that $\lambda_{{tn}_i}$ falls into the region $[0.2, 1.4]$ for almost all samples. For completeness, we summarize the training of MABCQ in Algorithm~\ref{alg:1}. 

\begin{table}
	\setlength{\extrarowheight}{3pt}
	\centering
	\caption{Transition probabilities and expected returns calculated in the dataset using only $\lambda_{{vd}}$.}
	\label{tab:1}
	\begin{small}
		\begin{tabular}{cccc}
			\multicolumn{1}{l}{}                                                    & action                                                              & transition                                  & expected return                            \\ \cline{2-4} 
			\multicolumn{1}{c|}{{\color[HTML]{9A0000} }}                            & \multicolumn{1}{c|}{{\color[HTML]{9A0000} }}                        & \multicolumn{1}{c|}{$p(1 | {\color[HTML]{9A0000} a_1}) = 0.12$} & \multicolumn{1}{c|}{}                      \\ \cline{3-3}
			\multicolumn{1}{c|}{{\color[HTML]{9A0000} }}                            & \multicolumn{1}{c|}{\multirow{-2}{*}{{\color[HTML]{9A0000} $a_1$}}} & \multicolumn{1}{c|}{$p(5 | {\color[HTML]{9A0000} a_1}) = 0.88$} & \multicolumn{1}{c|}{\multirow{-2}{*}{4.52}} \\ \cline{2-4} 
			\multicolumn{1}{c|}{{\color[HTML]{9A0000} }}                            & \multicolumn{1}{c|}{{\color[HTML]{9A0000} }}                        & \multicolumn{1}{c|}{$p(6 | {\color[HTML]{9A0000} a_2} ) = 0.8$} & \multicolumn{1}{c|}{}                      \\ \cline{3-3}
			\multicolumn{1}{c|}{\multirow{-4}{*}{\rotatebox{90}{\color[HTML]{9A0000} Agent $1$}}} & \multicolumn{1}{c|}{\multirow{-2}{*}{{\color[HTML]{9A0000} $a_2$}}} & \multicolumn{1}{c|}{$p(1 | {\color[HTML]{9A0000} a_2}) = 0.2$} & \multicolumn{1}{c|}{\multirow{-2}{*}{5}} \\ \cline{2-4} 
		\end{tabular}
		\begin{tabular}{cccc}
			\multicolumn{1}{l}{}                                                    & action                                                              & transition                                 & expected return                            \\ \cline{2-4} 
			\multicolumn{1}{c|}{{\color[HTML]{00009B} }}                            & \multicolumn{1}{c|}{{\color[HTML]{00009B} }}                        & \multicolumn{1}{c|}{$p(1 | {\color[HTML]{00009B} a_1}) = 0.4$} & \multicolumn{1}{c|}{}                      \\ \cline{3-3}
			\multicolumn{1}{c|}{{\color[HTML]{00009B} }}                            & \multicolumn{1}{c|}{\multirow{-2}{*}{{\color[HTML]{00009B} $a_1$}}} & \multicolumn{1}{c|}{$p(6 | {\color[HTML]{00009B} a_1}) = 0.6$} & \multicolumn{1}{c|}{\multirow{-2}{*}{4}} \\ \cline{2-4} 
			\multicolumn{1}{c|}{{\color[HTML]{00009B} }}                            & \multicolumn{1}{c|}{{\color[HTML]{00009B} }}                        & \multicolumn{1}{c|}{$p(5 | {\color[HTML]{00009B} a_2}) = 0.95$} & \multicolumn{1}{c|}{}                      \\ \cline{3-3}
			\multicolumn{1}{c|}{\multirow{-4}{*}{\rotatebox{90}{\color[HTML]{00009B} Agent $2$}}} & \multicolumn{1}{c|}{\multirow{-2}{*}{{\color[HTML]{00009B} $a_2$}}} & \multicolumn{1}{c|}{$p(1 | {\color[HTML]{00009B} a_2}) = 0.05$} & \multicolumn{1}{c|}{\multirow{-2}{*}{4.8}} \\ \cline{2-4} 
		\end{tabular}
	\end{small}
\end{table}

\begin{table}
	\setlength{\extrarowheight}{3pt}
	\centering
	\caption{Transition probabilities and expected returns calculated in the dataset using $\lambda_{{tn}}$ and $\lambda_{{vd}}$.}
	\label{tab:2}
	\begin{small}
		\begin{tabular}{cccc}
			\multicolumn{1}{l}{}                                                    & action                                                              & transition                                  & expected return                            \\ \cline{2-4} 
			\multicolumn{1}{c|}{{\color[HTML]{9A0000} }}                            & \multicolumn{1}{c|}{{\color[HTML]{9A0000} }}                        & \multicolumn{1}{c|}{$p(1 | {\color[HTML]{9A0000} a_1}) = 0.17$} & \multicolumn{1}{c|}{}                      \\ \cline{3-3}
			\multicolumn{1}{c|}{{\color[HTML]{9A0000} }}                            & \multicolumn{1}{c|}{\multirow{-2}{*}{{\color[HTML]{9A0000} $a_1$}}} & \multicolumn{1}{c|}{$p(5 | {\color[HTML]{9A0000} a_1}) = 0.83$} & \multicolumn{1}{c|}{\multirow{-2}{*}{4.33}} \\ \cline{2-4} 
			\multicolumn{1}{c|}{{\color[HTML]{9A0000} }}                            & \multicolumn{1}{c|}{{\color[HTML]{9A0000} }}                        & \multicolumn{1}{c|}{$p(6 | {\color[HTML]{9A0000} a_2}) = 0.86$} & \multicolumn{1}{c|}{}                      \\ \cline{3-3}
			\multicolumn{1}{c|}{\multirow{-4}{*}{\rotatebox{90}{\color[HTML]{9A0000} Agent $1$}}} & \multicolumn{1}{c|}{\multirow{-2}{*}{{\color[HTML]{9A0000} $a_2$}}} & \multicolumn{1}{c|}{$p(1 | {\color[HTML]{9A0000} a_2}) = 0.14$} & \multicolumn{1}{c|}{\multirow{-2}{*}{5.29}} \\ \cline{2-4} 
		\end{tabular}
		\begin{tabular}{cccc}
			\multicolumn{1}{l}{}                                                    & action                                                              & transition                                 & expected return                            \\ \cline{2-4} 
			\multicolumn{1}{c|}{{\color[HTML]{00009B} }}                            & \multicolumn{1}{c|}{{\color[HTML]{00009B} }}                        & \multicolumn{1}{c|}{$p(1 | {\color[HTML]{00009B} a_1}) = 0.14$} & \multicolumn{1}{c|}{}                      \\ \cline{3-3}
			\multicolumn{1}{c|}{{\color[HTML]{00009B} }}                            & \multicolumn{1}{c|}{\multirow{-2}{*}{{\color[HTML]{00009B} $a_1$}}} & \multicolumn{1}{c|}{$p(6 | {\color[HTML]{00009B} a_1}) = 0.86$} & \multicolumn{1}{c|}{\multirow{-2}{*}{5.29}} \\ \cline{2-4} 
			\multicolumn{1}{c|}{{\color[HTML]{00009B} }}                            & \multicolumn{1}{c|}{{\color[HTML]{00009B} }}                        & \multicolumn{1}{c|}{$p(5 | {\color[HTML]{00009B} a_2}) = 0.83$} & \multicolumn{1}{c|}{}                      \\ \cline{3-3}
			\multicolumn{1}{c|}{\multirow{-4}{*}{\rotatebox{90}{\color[HTML]{00009B} Agent $2$}}} & \multicolumn{1}{c|}{\multirow{-2}{*}{{\color[HTML]{00009B} $a_2$}}} & \multicolumn{1}{c|}{$p(1 | {\color[HTML]{00009B} a_2}) = 0.17$} & \multicolumn{1}{c|}{\multirow{-2}{*}{4.33}} \\ \cline{2-4} 
		\end{tabular}
	\end{small}
\end{table}

\begin{table*}
	\renewcommand{\arraystretch}{1.1}
	\centering
	\caption{Normalized scores of MABCQ and the baselines.}
	\label{tab:reward}
	\begin{small}
		\begin{tabular}{ccccccc}
			\toprule
			& MABCQ           & BCQ w/ $\lambda_{vd}$ & BCQ w/ $\lambda_{tn}$ & BCQ             & DDPG            & Behavior       \\ \midrule
			HalfCheetah & $\bm{17.6} \pm 3.3$  & $13.3 \pm 4.8$           & $13.4 \pm 6.5$           & $13.4 \pm 5.2$  & $-3.1 \pm 3.6$  & $11.3 \pm 2.8$ \\
			Walker      & $\bm{54.4} \pm 5.6$  & $50.1 \pm 11.0$          & $41.2 \pm 17.8$          & $28.8 \pm 14.4$ & $1.7 \pm 0.9$   & $10.0 \pm 0.8$ \\
			Hopper      & $\bm{43.1} \pm 14.2$ & $34.1 \pm 8.2$           & $19.8 \pm 8.7$           & $18.0 \pm 3.4$  & $9.0 \pm 15.3$  & $10.7 \pm 2.3$ \\
			Ant         & $\bm{60.5} \pm 3.6$  & $\bm{59.7} \pm 4.9$           & $\bm{62.9} \pm 2.1$      & $51.5 \pm 12.7$ & $-48.4 \pm 1.3$ & $19.3 \pm 4.5$\\
			\bottomrule
		\end{tabular}
	\end{small}
\end{table*}

\section{Related Work}

\subsection{Off-policy MARL} Many off-policy MARL methods have been proposed for learning to solve cooperative tasks in an online manner. Policy-based methods \cite{lowe2017multi,iqbal2019actor,zhang2021fop, su2021divergence, wang2023more} extend actor-critic into multi-agent cases. Value factorization methods \cite{sunehag2018value,rashid2018qmix,son2019qtran,wang2020qplex} decompose the joint value function into individual value functions. All these methods follow CTDE, where the information of all agents can be accessed in a centralized way during training. Unlike these studies, we consider decentralized settings where global information is not available. 
For decentralized learning, the key challenge is the obsolete experiences in the replay buffer, which is considered in Fingerprints \cite{foerster2017stabilising}, Lenient-DQN \cite{palmer2018lenient}, and concurrent experience replay \cite{omidshafiei2017deep}. However, these methods require additional information, \textit{e.g.}, training iteration number and exploration rate, which are often not provided by the offline dataset. 

\subsection{Offline RL} 
Offline RL requires the agent to learn from a fixed batch of data consisting of single-step transitions, without exploration. 
Most offline RL methods consider the out-of-distribution action \cite{levine2020offline} as the fundamental challenge, which is the main cause of the extrapolation error \cite{fujimoto2019off} in value estimate in the single-agent environment. To minimize the extrapolation error, some recent methods introduce constraints to enforce the learned policy to be close to the behavior policy, which can be direct action constraint \cite{fujimoto2019off}, kernel MMD \cite{kumar2019stabilizing}, Wasserstein distance \cite{wu2019behavior}, KL divergence \cite{peng2019advantage}, or $l2$ distance \cite{fujimoto2021minimalist,pan2021plan}. Some methods train a Q-function pessimistic to out-of-distribution actions to avoid overestimation by adding a reward penalty quantified by the learned environment model \cite{yu2020mopo}, by minimizing the Q-values of out-of-distribution actions \cite{kumar2020conservative,yu2021combo}, by weighting the update of Q-function via Monte Carlo dropout \cite{wu2021uncertainty}, or by explicitly assigning and training pseudo Q-values for out-of-distribution actions \cite{lyu2022mildly}. Our framework can be built on these methods.

MAICQ \cite{yang2021believe} studies offline MARL in the CTDE setting, which requires the joint actions of all agents in the dataset and cannot be applied to decentralized settings where datasets contain only individual actions. 
\textit{All the methods aforementioned do not consider the extrapolation error introduced by the transition bias, which is a fatal problem in offline decentralized MARL. }

\section{Experiments}

We evaluate our framework in both fully and partially observable tasks. In each task, we build offline dataset $\mathcal{B}_i$ for each agent $i$, which \textit{does not contain actions of other agents}. We will give the details about the collection of each offline dataset. Our method and baselines have the same neural network architectures and hyperparameters, which are available in Appendix. All the models are trained for five runs with different random seeds, and the results are presented in terms of mean and std.

\subsection{Matrix Game}

We perform MABCQ on the matrix game in Table~\ref{tab:1step_game}. As shown in Table~\ref{tab:1}, if we only use $\lambda_{vd}$ without considering \textit{transition normalization}, since the transition probabilities of high-value next states have been increased, for agent $1$ the value of ${\color[HTML]{9A0000} a_2}$ becomes higher than that of ${\color[HTML]{9A0000} a_1}$. However, due to the unbalanced action distribution of agent $1$, the initial transition probabilities of agent $2$ are extremely unbalanced. With $\lambda_{vd}$, agent $2$ still underestimates the value of ${\color[HTML]{00009B} a_1}$ and learns the action ${\color[HTML]{00009B} a_2}$. The agents arrive at the joint action $({\color[HTML]{9A0000} a_2},{\color[HTML]{00009B} a_2})$, which is a worse solution than the initial one (Table~\ref{tab:step_game_return}). Further, by normalizing the transition probabilities by $\lambda_{tn}$, the agents can learn the optimal solution $({\color[HTML]{9A0000} a_2},{\color[HTML]{00009B} a_1})$ and build the consensus about the values of learned actions, as shown in Table~\ref{tab:2}.

\begin{table}
	\renewcommand{\arraystretch}{1.1}
	\centering
	\caption{Mean difference in value estimates among agents during training. It is shown that \textit{transition normalization} indeed reduces the difference in value estimates. }
	\label{tab:v}
	\begin{small}
		\begin{tabular}{ccc}
			\toprule
			value difference & MABCQ               & BCQ w/ $\lambda_{vd}$              \\\midrule
			HalfCheetah & $\bm{44.4}\pm3.4$       & $411.7\pm72.4$ \\
			Walker      & $\bm{28.2}\pm2.8$   & $38.7\pm6.9$      \\
			Hopper      & $\bm{24.2}\pm0.8$   & $25.4\pm1.3$       \\
			Ant         & $\bm{60.8}\pm2.9$ & $67.0\pm3.1$      \\
			\bottomrule
		\end{tabular}
	\end{small}	
\end{table}

\begin{table}
	\renewcommand{\arraystretch}{1.1}
	\centering
	\caption{Extrapolation errors. It is shown that our framework can decrease the extrapolation error.}
	\label{tab:e}
	\begin{small}
		\begin{tabular}{ccc}
			\toprule
			extrapolation error & MABCQ               & BCQ                \\\midrule
			HalfCheetah & $98.4\pm31.3$       & $97.2\pm29.1$ \\
			Walker      & $\bm{55.0}\pm9.6$   & $91.5\pm35.4$      \\
			Hopper      & $\bm{28.1}\pm3.4$   & $65.8\pm6.4$       \\
			Ant         & $\bm{180.2}\pm22.2$ & $231.3\pm47$      \\
			\bottomrule
		\end{tabular}
	\end{small}	
\end{table}

\begin{table*}[th]
	\renewcommand{\arraystretch}{1}
	\centering
	\caption{Rewards on SMAC datasets.}
	\label{tab:star}
	\begin{small}
		\begin{tabular}{cccccc}
			\toprule
			&            & MABCQ            & BCQ              & MACQL & CQL \\\midrule
			\multirow{4}{*}{random} & 3m         & $\bm{4.0} \pm 1.1$  & $0 \pm 0$        & $\bm{13.5} \pm 1.5$ & $9.3 \pm 3.0$ \\
			& 8m         & $\bm{4.5} \pm 0.9$  & $0.8 \pm 0.2$  & $\bm{8.2} \pm 1.1$&$6.6 \pm 1.2$\\
			& 3s\_vs\_3z & $\bm{8.1}\pm0.7$    & $0 \pm 0$        & $10.2 \pm 0.8$ & $10.1 \pm 0.4$\\
			& 3s\_vs\_4z & $\bm{4.1}\pm1.5$    & $0 \pm 0$        & $5.7 \pm 0.6$ & $6.7 \pm 0.2$ \\ \midrule
			\multirow{4}{*}{medium} & 3m         & $8.9 \pm 1.3$    & $7.8 \pm 0.4$      & $\bm{15.1} \pm 1.8$ & $13.8 \pm 1.4$\\
			& 8m         & $\bm{7.6} \pm 0.8$  & $4.5 \pm 1.2$       & $\bm{14.5} \pm 1.5$ & $12.4 \pm 0.6$\\
			& 3s\_vs\_3z & $\bm{8.7} \pm1.1$   & $3.9 \pm 0.6$       & $9.3 \pm 0.9$ & $8.9 \pm 0.6$\\
			& 3s\_vs\_4z & $\bm{4.3} \pm 0.5$  & $0 \pm 0$       & $6.1 \pm 0.7$ & $6.8 \pm 1.8$\\\midrule
			\multirow{4}{*}{replay} & 3m         &  $13.2 \pm 0.2$ & $12.7 \pm 0.7$  & $13.8 \pm 0.4$&$13.5 \pm 0.6$\\
			& 8m         & $\bm{15.2} \pm 1.0$  & $14.3 \pm 0.9$  & $\bm{17.9} \pm 0.4$& $16.3 \pm 0.4$\\
			& 3s\_vs\_3z & $19.4 \pm 0.4$  & $19.8 \pm 0.3$  & $20.0 \pm 0.0$ & $20.0 \pm 0.0$ \\
			& 3s\_vs\_4z & $5.3 \pm 0.6$  & $5.3 \pm 0.9$  & $\bm{5.9} \pm 0.3$ &$5.2 \pm 0.7$\\\midrule
			\multirow{4}{*}{expert} & 3m         &  $18.8 \pm 0.7$ &  $18.3 \pm 1.1$ & $18.9 \pm 0.5$&$19.1 \pm 0.5$\\
			& 8m         &  $17.0 \pm 0.8$ &  $17.5 \pm 1.1$ &$18.5 \pm 1.2$ &$18.3 \pm 1.1$\\
			& 3s\_vs\_3z & $19.1 \pm 0.5$  &$19.0 \pm 0.6$   &$19.2 \pm 0.6$ &$19.1 \pm 0.6$\\
			& 3s\_vs\_4z &  $5.6 \pm 0.9$ &   $5.4 \pm 1.1$& $6.8 \pm 0.7$&$6.5 \pm 0.9$\\
			\bottomrule
		\end{tabular}
	\end{small}
\end{table*}
\subsection{Multi-Agent Mujoco}

To evaluate MABCQ in high-dimensional complex environments, we adopt multi-agent mujoco \cite{de2020deep}, where each agent independently controls one or some joints of the robot and can get the state \cite{kuba2022trust} and reward of the robot. The task illustration and the collection of offline datasets are given in Appendix. 

\textbf{Baselines.} 
We compare MABCQ against
\begin{itemize}
	\setlength\itemsep{0em}
	\item {BCQ w/ ${\lambda_{vd}}$}. Using $\lambda_{vd}$ alone on BCQ.
	\item {BCQ w/ ${\lambda_{tn}}$}. Using $\lambda_{tn}$ alone on BCQ.
	\item {BCQ}. Removing both $\lambda_{tn}$ and $\lambda_{vd}$ from MABCQ.
	\item {DDPG} \cite{lillicrap2016continuous}. Each agent $i$ is trained using independent DDPG on the offline $\mathcal{B}_i$ without action constraint and transition probability modification.
	\item {Behavior}. Each agent $i$ takes the action generated from the VAE $G_i^1$.
\end{itemize}

\begin{figure*}[!ht]
	\centering
	\begin{subfigure}{.24\linewidth}
		\setlength{\abovecaptionskip}{1pt}
		\includegraphics[width=1\linewidth]{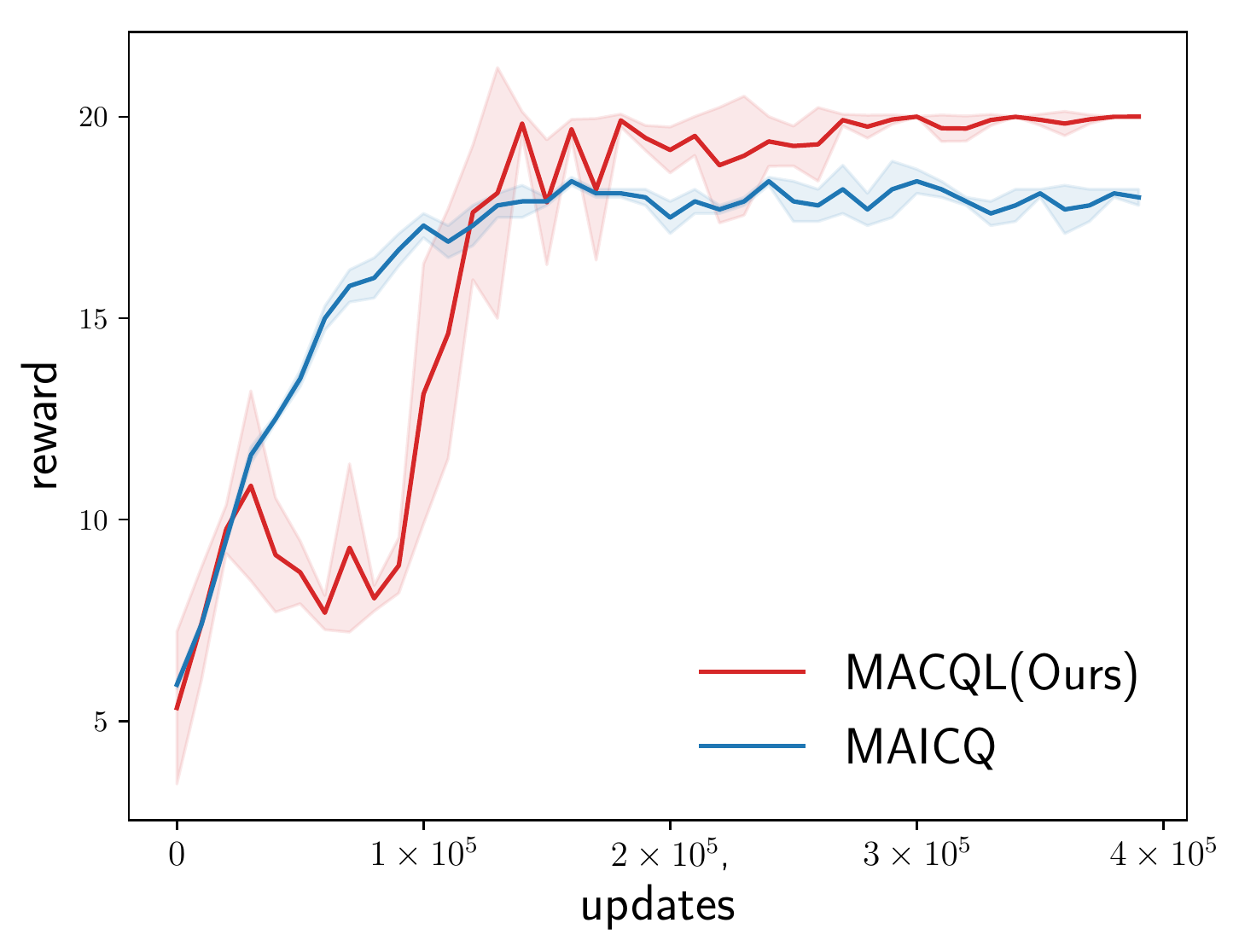}
		\caption{3s\_vs\_3z}
		\label{fig:smac1}
	\end{subfigure}
	\begin{subfigure}{.24\linewidth}
		\setlength{\abovecaptionskip}{1pt}
		\includegraphics[width=1\linewidth]{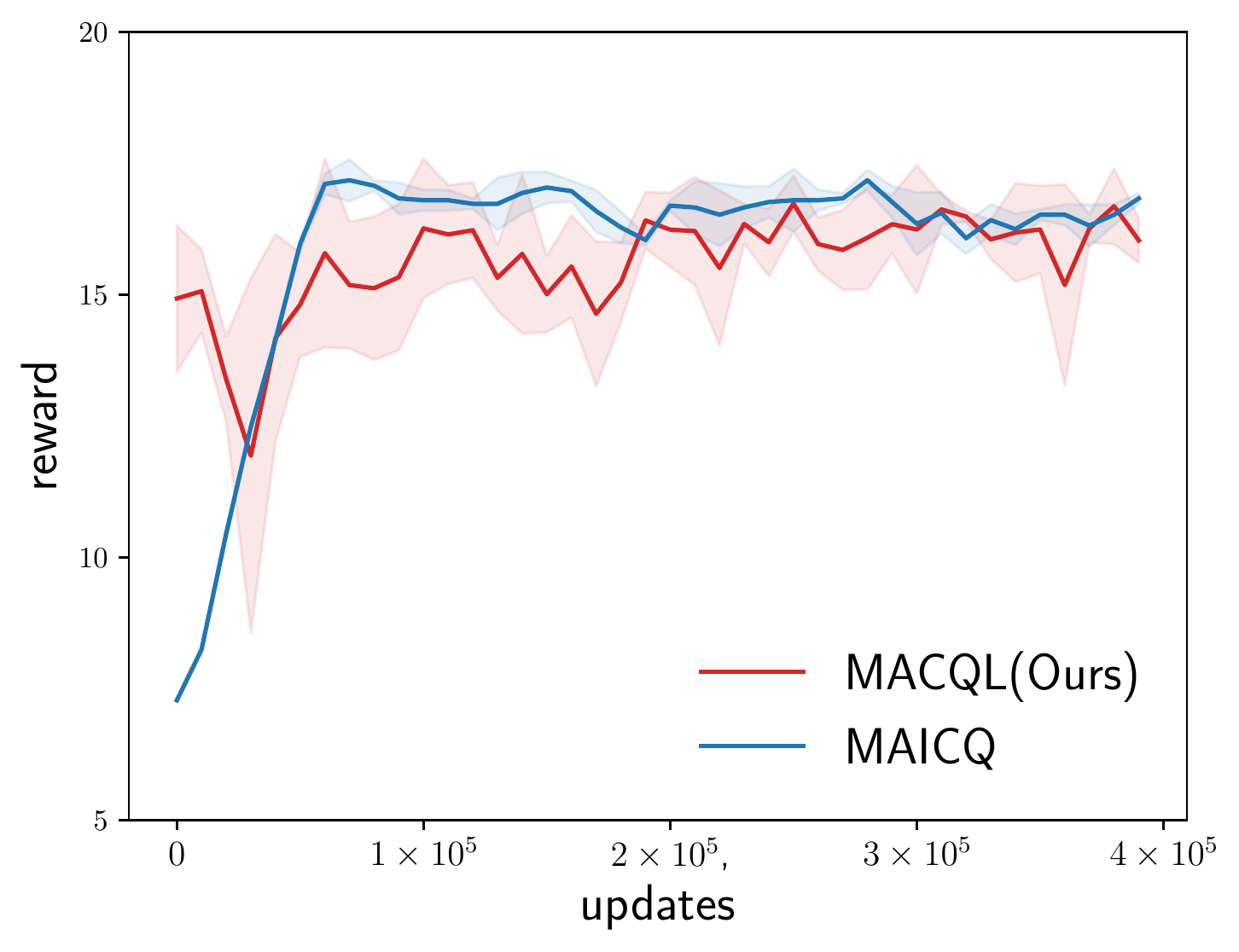}
		\caption{2s3z}
		\label{fig:smac2}
	\end{subfigure}
	\begin{subfigure}{.234\linewidth}
		\setlength{\abovecaptionskip}{1pt}
		\includegraphics[width=1\linewidth]{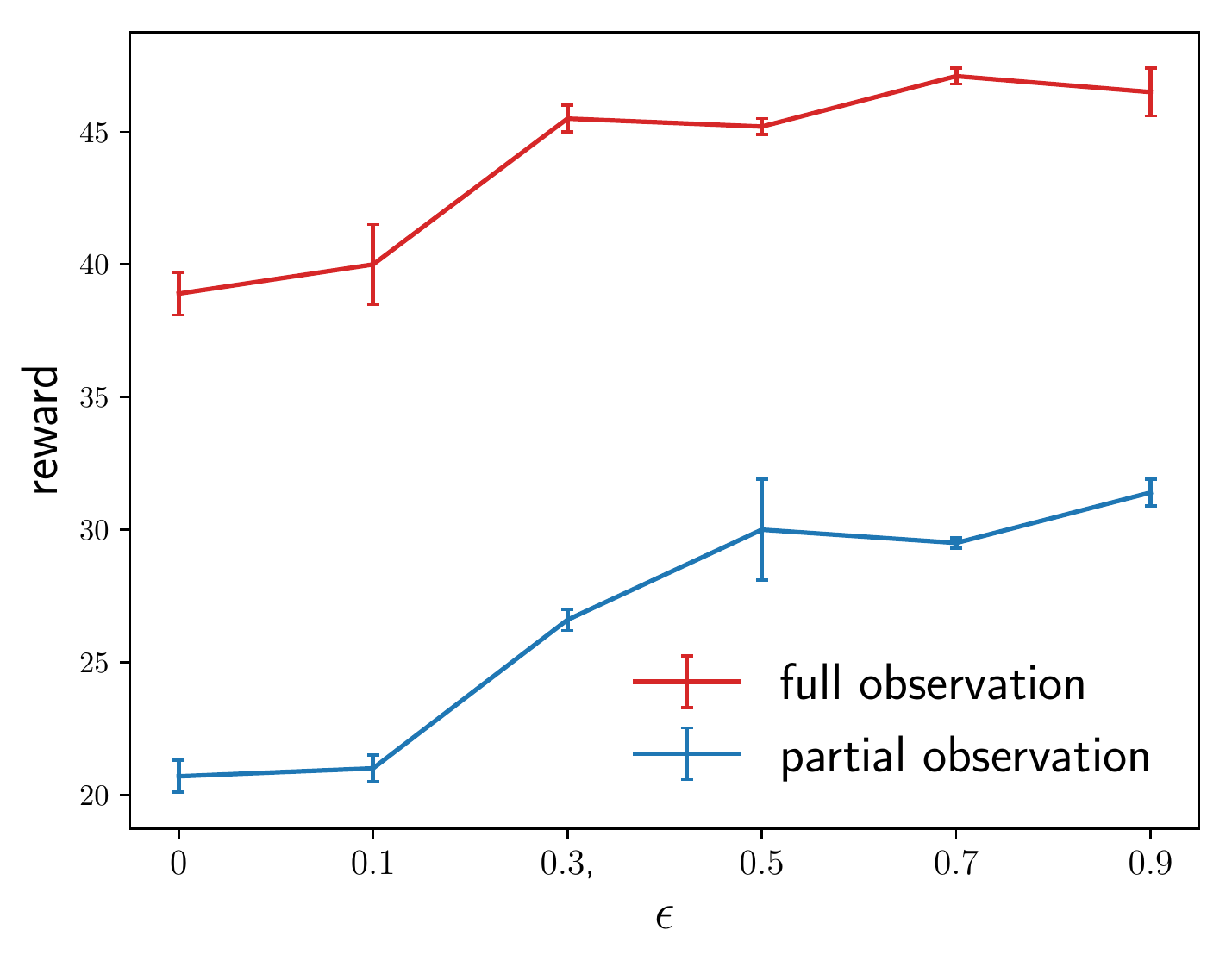}
		\caption{DG}
		\label{fig:epsilon}
	\end{subfigure}
	\begin{subfigure}{.24\linewidth}
		\setlength{\abovecaptionskip}{1pt}
		\includegraphics[width=1\linewidth]{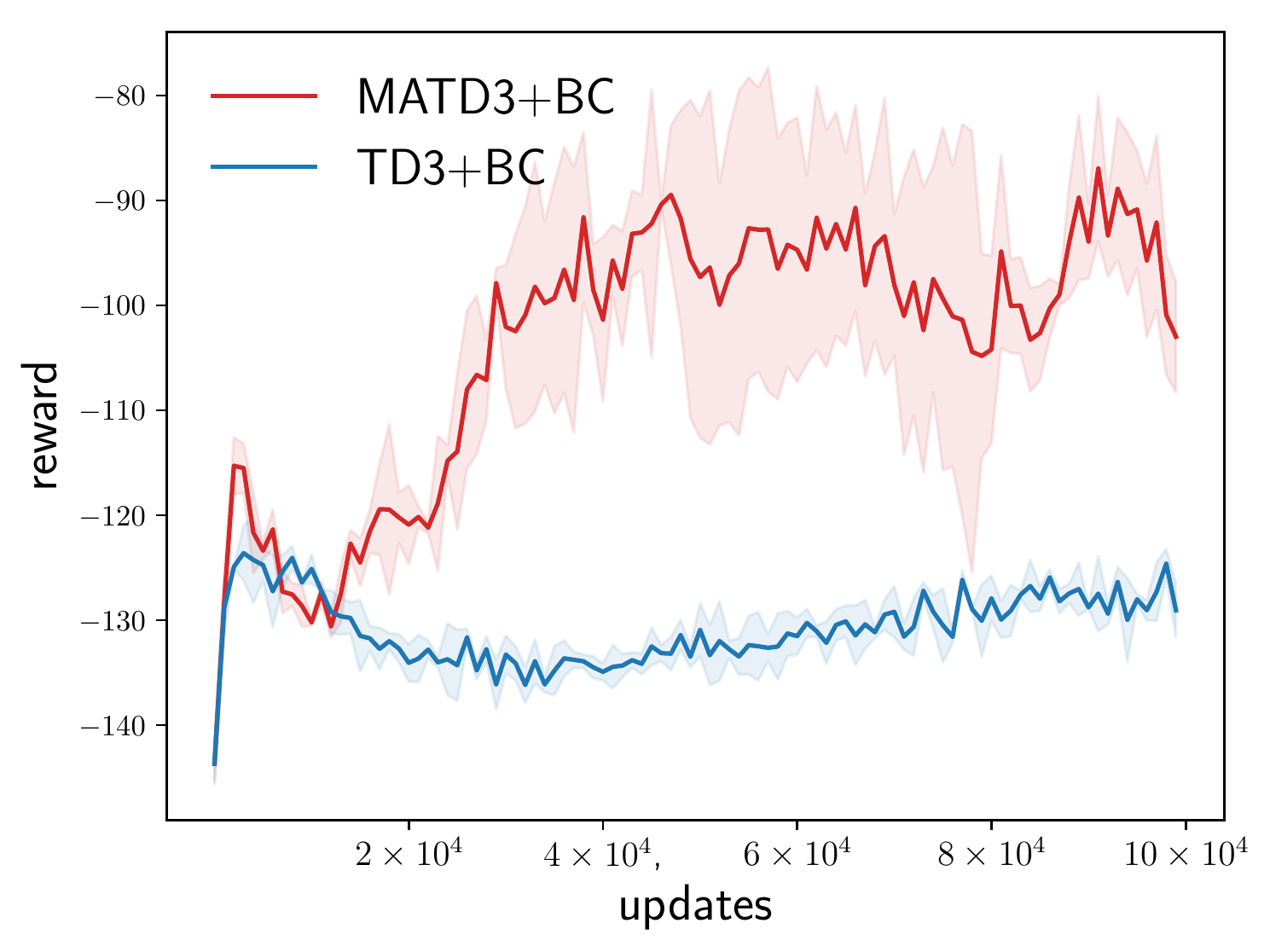}
		\caption{CN}
		\label{fig:cn}
	\end{subfigure}
	\caption{(a) and (b): Comparison with MAICQ on two SMAC datasets \cite{yang2021believe}. (c): Performance of MATD3+BC with different $\epsilon$ on DG datasets. (d): Learning curves on CN datasets.}
\end{figure*}

\textbf{Ablation.}
Table~\ref{tab:reward} shows the normalized scores \cite{fu2020d4rl} of all the methods in the four tasks. Without action constraint, DDPG severely suffers from the large extrapolation error and hardly learns. BCQ outperforms the behavior policies but only arrives at mediocre performance. Using $\lambda_{tn}$ alone does not always improve the performance, \textit{e.g.}, in HalfCheetah and Hopper. This is because $\lambda_{tn}$ makes transition probabilities be uniform, which can be far from the ones in execution, leading to large extrapolation errors. In Ant, BCQ w/ $\lambda_{tn}$ outperforms BCQ, which is attributed to the value consensus built by the normalized transition probabilities. By optimistically increasing the transition probabilities of high-value next states, $\lambda_{vd}$ mitigates the underestimation of potential good actions and thus boosts the performance. MABCQ combines the advantages of value deviation and transition normalization and outperforms other baselines.

\textbf{Consensus on value estimates.} 
To verify that \textit{transition normalization} can decrease the difference in value estimates among agents, we uniformly sample a subset from the union of all agents' states and calculate the difference in value estimates, $\max_i V^*_i - \min_i V^*_i$, on this subset. The mean differences during training are illustrated in Table~\ref{tab:v}. The $\max_i V^*_i - \min_i V^*_i$ of MABCQ is indeed lower than that of BCQ w/ $\lambda_{vd}$. If there is a consensus among agents about which states are high-value, the agents will select the actions that most likely lead to the common high-value states. This promotes the coordination of policies and helps MABCQ outperform BCQ w/ $\lambda_{vd}$.


\textbf{Extrapolation error.}
In Table~\ref{tab:e}, we present the extrapolation errors of MABCQ and BCQ, measured by $|\frac{1}{N}\sum_{i}Q_i(s,a_i)-G|$, where $G$ is the true return evaluated by Monte Carlo.  Although MABCQ greatly outperforms BCQ (\textit{i.e.}, much higher return), it still achieves much smaller extrapolation errors than BCQ in Walker, Hopper, and Ant. This empirically verifies the claim that our method can decrease the extrapolation error.

\subsection{SMAC}

We also investigate the proposed framework on SMAC \cite{samvelyan19smac} tasks, including 3m, 8m, 3s\_vs\_3z, and 3s\_vs\_4z. We build random datasets that are generated by uniform policies, medium datasets that are generated by mixed medium and uniform policies, replay datasets that are collected in the training process of QMIX \cite{rashid2018qmix}, and expert datasets that are generated by expert policies trained by QMIX. Each dataset contains $1 \times 10^4$ episodes. We also build our framework on CQL \cite{kumar2020conservative}, as MACQL. As shown in Table~\ref{tab:star}. MABCQ and MACQL achieve great performance improvement compared with the baselines, especially in random and medium datasets, where the transition dynamics in execution are much different from the ones in offline datasets. In expert datasets, since the behavior policies are much deterministic, offline RL methods avoid selecting out-of-distribution actions and thus degenerate to behavior cloning. Therefore, all the methods perform similarly.

\textbf{Offline CTDE settings.} 
Although offline CTDE method \cite{yang2021believe} does not fit our offline decentralized setting, our framework can work in offline CTDE datasets. To verify, we select two replay datasets (jointly collected) in MAICQ \cite{yang2021believe}, split them into individual datasets, and test MACQL on them. As shown in Figures~\ref{fig:smac1} and~\ref{fig:smac2}, our decentralized method can obtain competitive performance compared with the centralized method, MAICQ. 


\begin{table}
	\centering
	\caption{Rewards on DG datasets.}
	\label{tab:mpe}
	\begin{small}
		\begin{tabular}{ccc}
			\toprule
			& MATD3+BC & TD3+BC \\\midrule
			full observation  &  $\bm{46.5} \pm 0.9$     &    $38.9 \pm 0.8$    \\
			partial observation & $\bm{31.4} \pm 0.6$  &   $20.7 \pm 0.6$    \\
			\bottomrule
		\end{tabular}
	\end{small}
\end{table}

\begin{table*}[t]
	\renewcommand{\arraystretch}{1.1}
	\centering
	\caption{Normalized scores on D4RL MuJoCo datasets.}
	\label{tab:d4rl}
	\begin{small}
		\begin{tabular}{ccccc}
			\toprule
			&             & MATD3+BC            & TD3+BC          & TD3+BC (single) \\ \midrule
			\multirow{3}{*}{random}        & halfcheetah & $\bm{14.3} \pm 2.9$ & $12.9 \pm 2.6$  & $10.2 \pm 1.3$  \\
			& hopper      & $10.3 \pm 0.6$      & $10.4 \pm 0.8$  & $11.0 \pm 0.1$  \\
			& walker2d    & $4.5 \pm 3.4$       & $3.7 \pm 3.0$   & $1.4 \pm 1.6$   \\ \midrule
			\multirow{3}{*}{medium}        & halfcheetah &  $\bm{41.5} \pm 2.4$         &    $40.4 \pm 2.4$     & $42.8 \pm 0.3$  \\
			& hopper      & $97.1 \pm 1.7$      & $98.7 \pm 1.2$  & $99.5 \pm 1.0$  \\
			& walker2d    & $\bm{82.0} \pm 5.1$ & $72.3 \pm 4.1$  & $79.7 \pm 1.8$  \\ \midrule
			\multirow{3}{*}{replay}        & halfcheetah &      $40.0 \pm 3.2$               & $39.3 \pm 2.8$  & $43.3 \pm 0.5$  \\
			& hopper      & $\bm{28.4} \pm 4.1$ & $25.5 \pm 3.3$  & $31.4 \pm 3.0$  \\
			& walker2d    &      $21.0 \pm 2.9$          &    $21.3 \pm 1.5$          & $25.2 \pm 5.1$  \\ \midrule
			\multirow{3}{*}{medium-expert} & halfcheetah & $96.2 \pm 5.5$      & $95.3 \pm 6.5$  & $97.9 \pm 4.4$  \\
			& hopper      & $112.0 \pm 0.8$     & $112.3 \pm 0.9$ & $112.2 \pm 0.2$ \\
			& walker2d    &   $88.9 \pm 17.4$   & $82.0 \pm 10.8$ & $105.7 \pm 2.7$\\
			\bottomrule
		\end{tabular}
	\end{small}
\end{table*}

\subsection{MPE}
\begin{figure}
	\centering
	\includegraphics[width=.6\linewidth]{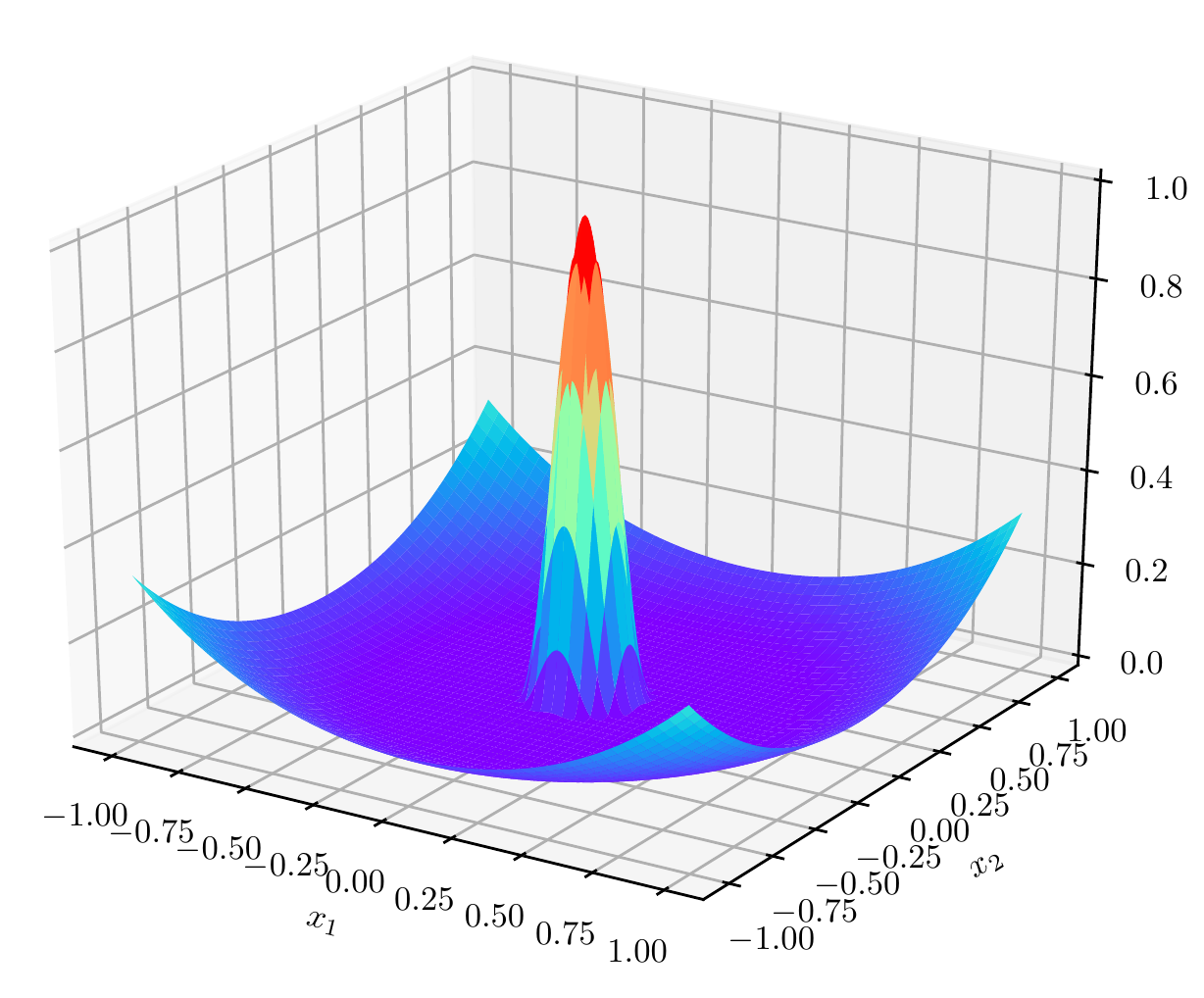}
	\caption{Visualization of reward function in Differential Game.}
	\label{fig:mpe_reward}
\end{figure}

We additionally evaluate our framework in an MPE-based \cite{lowe2017multi} Differential Game (DG), where the transition bias greatly affects the performance. Two agents can move in the range $[-1,1]$. The action is the speed, which is in the range $[-0.1,0.1]$. Define $l = \sqrt{x_1^2+x_2^2}$, where $x_1$ and $x_2$ are the positions of the two agents, respectively. The shared reward is set as
\begin{small}
	\begin{equation}
	\notag
	r = \begin{cases}
	0.5 \times (\cos(15 \times l)+1) & \text{ if } l<0.2 \\ 
	0 & \text{ if } 0.2\leq l\leq 0.6 \\ 
	0.5 \times (l-0.6)^2 & \text{ if } l>0.6. 
	\end{cases}
	\end{equation}
\end{small}
The visualization of reward function is shown in Figure~\ref{fig:mpe_reward}. 

\textbf{Partial observation vs full observation.}
The offline datasets are collected by uniform random policies, containing $1 \times 10^6$ transitions. In the full observation setting, the dataset of each agent contains the positions of both agents. In the partial observation setting, the dataset of each agent only contains its own position. In both settings, the datasets do not contain the actions of the other agent.
We add $\lambda_{vd}$ and $\lambda_{tn}$ to TD3+BC \cite{fujimoto2021minimalist}, as MATD3+BC. As shown in Table~\ref{tab:mpe}, MATD3+BC obtains the substantial improvement in both full and partial observation settings.

\textbf{Hyperparameter $\bm{\epsilon}$.}
The optimism level $\epsilon$ controls the strength of value deviation. If $\epsilon$ is too small, value deviation has weak effects on the objective function. On the other hand, if $\epsilon$ is too large, the agent will be overoptimistic about other agents' learned policies.  Figure~\ref{fig:epsilon} shows that the performance of MATD3+BC with different $\epsilon$, which verifies our framework is robust to $\epsilon$.

We test MATD3+BC on Cooperative Navigation in MPE, where $4$ agents learn to cover $4$ landmarks. The reward is $-\text{sum}(\mathrm{distance}_j)$, where $\mathrm{distance}_j$ is the distance from landmark $j$ to the closest agent. The offline datasets are collected by uniform random policies, containing $1 \times 10^6$ transitions. Figure~\ref{fig:cn} shows our framework significantly outperforms the baseline.

\subsection{Additional Results}

We also split the D4RL \cite{fu2020d4rl} Mujoco datasets into decentralized multi-agent datasets, and test MATD3+BC on them. The results are summarized in Table~\ref{tab:d4rl}. We find the results of decentralized methods, where the joints are controlled by different agents, are very close to the results of single-agent method, TD3+BC (single), where a single agent controls all joints, which could be seen as the ``upper-bound'' of the decentralized methods. That is the reason that our method does not bring significant improvement in these tasks. However, MATD3+BC still outperforms TD3+BC on several tasks, \textit{e.g.}, halfcheetah-random and walker2d-medium.

\begin{table}
	\renewcommand{\arraystretch}{1.1}
	\centering
	\setlength{\abovecaptionskip}{0cm}
	\caption{Average time taken by one update.}
	\label{tab:time}
	\begin{small}
		\begin{tabular}{cccc}
			\toprule
			MABCQ    & BCQ &  MATD3+BC &   TD3+BC    \\\midrule
			$18$ ms       & $10$ ms  & $4$ ms       & $3$ ms\\
			\bottomrule
		\end{tabular}
	\end{small}
\end{table}

To demonstrate the computation efficiency of our method, in Table~\ref{tab:time}, we record the average time taken by one update in Halfcheetah. The experiments are carried out on Intel i7-8700 CPU and NVIDIA GTX 1080Ti GPU. Since $\lambda_{vd}$ and $\lambda_{tn}$ could be calculated from the sampled experience without actually going through all next states, our framework additionally needs only two forward passes for computing $\lambda_{vd}$ and $\lambda_{tn}$ in the update. Since the value computation is very efficient in TD3+BC, our framework is also efficient on it. 

\section{Conclusion}

We propose a framework for offline decentralized multi-agent reinforcement learning, to overcome the mismatch between transition dynamics. The framework can be instantiated on many offline RL methods. Theoretically, we show that under the purposely controlled non-stationary transition dynamics, offline decentralized Q-learning converges to a unique fixed point. Empirically, the framework outperforms the baselines in a variety of multi-agent offline datasets.

\ack This work was supported by NSF China under grants 62250068 and 61872009. The authors would like to thank the anonymous reviewers for their valuable comments.

\bibliography{ecai}


\clearpage
\appendix

\section{Proofs} 
\label{appendix:proof}

\setcounter{proposition}{0}
\setcounter{Theorem}{0}

\begin{proposition}
	In episodic environments, if each agent $i$ performs Q-learning on $\mathcal {B}_i$, all agents will converge to the same $V^*$ if they have the same transition probability on any state where each agent $i$ acts the learned action $a_i^*$.
	\begin{proof} 
		Considering the two-agent case, we define $\delta(s)$ as the difference in the $V^*$.
		\begin{small}
			\begin{align*}
			\delta(s)&=V_{1}^{*}(s)-V_{2}^{*}(s)\\
			&=\sum_{s^{\prime}} P_{\mathcal {B}_1}\left(s^{\prime} | s, a_{1}^{*}\right)\left(r+\gamma V_{1}^{*}\left(s^{\prime}\right)\right)\\
			&\quad-\sum_{s^{\prime}} P_{\mathcal {B}_{2}}\left(s^{\prime} | s, a_{2}^{*}\right)\left(r+\gamma V_{2}^{*}\left(s^{\prime}\right)\right)\\
			&=\sum_{s^{\prime}} P_{\mathcal {B}_{1}}\left(s^{\prime} | s, a_{1}^{*}\right)\left(r+\gamma V_{2}^{*}\left(s^{\prime}\right)+\gamma V_{1}^{*}\left(s^{\prime}\right) -\gamma V_{2}^{*}\left(s^{\prime}\right)\right)  \\
			&\quad -\sum_{s^{\prime}} P_{\mathcal {B}_{2}}\left(s^{\prime} | s, a_{2}^{*}\right)\left(r+\gamma V_{2}^{*}\left(s^{\prime}\right)\right)\\
			&=\sum_{s^{\prime}} \left(P_{\mathcal {B}_{1}}\left(s^{\prime} | s, a_{1}^{*}\right)-P_{\mathcal {B}_{2}}\left(s^{\prime} | s, a_{2}^{*}\right)\right)\left(r+\gamma V_{2}^{*}\left(s^{\prime}\right)\right) \\
			&\quad+\gamma P_{\mathcal {B}_{1}}\left(s^{\prime} | s, a_{1}^{*}\right) \delta\left(s^{\prime}\right)
			\end{align*}
		\end{small}
		For the terminal state $s_{end}$, we have $\delta(s_{end}) = 0$. If $P_{\mathcal {B}_{1}}\left(s^{\prime} | s, a_{1}^{*}\right) = P_{\mathcal {B}_{2}}\left(s^{\prime} | s, a_{2}^{*}\right),\, \forall s' \in S$, recursively expanding the $\delta$ term, we arrive at $\delta(s) = 0 + \gamma0 + \gamma^20 + ... +0= 0$. We can easily show that it also holds in the $N$-agent case.
	\end{proof}
\end{proposition}

\begin{Theorem}
	Under the non-stationary transition probability $\hat{P}_{\mathcal {B}_i}$, the Bellman operator $\mathcal{T}$ is a contraction and converges to a unique fixed point when $\gamma < \frac{r_{\min}}{2r_{\max} - r_{\min}}$, if the reward is bounded by the positive region $[r_{\min},r_{\max}]$.
	
	\begin{proof} 
		We initialize the Q-value to be $\eta r_{\min}$, where $\eta$ denotes $\frac{1-\gamma^{T+1}}{1-\gamma}$. Since the reward is bounded by the positive region $[r_{\min},r_{\max}]$, the Q-value under the operator $\mathcal{T}$ is bounded to $[\eta r_{\min},\eta r_{\max}]$. Based on the definition of $\hat{P}_{\mathcal{B}_i}\left(s^{\prime} | s, a_i\right)$, it can be written as $\frac{V_i^{*}(s')}{\sum_{\hat{s}'}V_i^{*}({s}')}$, where $V_i^{*}({s}') = \max _{\hat{a}_i} Q_i\left(s^{\prime}, \hat{a}_i\right)$. Then, we have the following,
		\begin{small}
			\begin{align*}
			&\left\|\mathcal{T} Q_{i}^1-\mathcal{T} Q_{i}^2\right\|_{\infty}\\
			&=\max _{s, a_i} \left | \sum_{s' \in \mathcal{S}}\hat{P}^1_{\mathcal {B}_i}\left(s^{\prime} | s, a_i\right)\left[r+\gamma \max _{\hat{a}_i} Q_i^1\left(s^{\prime}, \hat{a}_i\right)\right]  \right.\\
			& \left. \quad - \sum_{s' \in \mathcal{S} }\hat{P}^2_{\mathcal {B}_i}\left(s^{\prime} | s, a_i\right)\left[r+\gamma \max _{\hat{a}_i} Q^2_i\left(s^{\prime}, \hat{a}_i\right)\right] \right |\\
			&= \max _{s, a_i}\gamma \left |\frac{\sum_{s' \in \mathcal{S} }(V_i^{*^1}(s'))^2}{\sum_{{s}'\in \mathcal{S}}V_i^{*^1}({s}')}-\frac{\sum_{s'\in \mathcal{S}}(V_i^{*^2}(s'))^2}{\sum_{{s}'\in \mathcal{S}}V_i^{*^2}({s}')} \right |\\
			&= \max _{s, a_i}\gamma \left |\frac{\sum_{s'\in \mathcal{S}}(V_i^{*^1}(s'))^2-(V_i^{*^2}(s'))^2}{\sum_{{s}'\in \mathcal{S}}V_i^{*^1}({s}')}\right.\\
			& \left. \quad -\sum_{s'\in \mathcal{S}}(V_i^{*^2}(s'))^2\left(\frac{1}{\sum_{{s}'\in \mathcal{S}}V_i^{*^2}({s}')} - \frac{1}{\sum_{{s}'\in \mathcal{S}}V_i^{*^1}({s}')} \right) \right|\\
			&= \max _{s, a_i}\gamma \left |\frac{\sum_{s'\in \mathcal{S}}(V_i^{*^1}(s')-V_i^{*^2}(s'))(V_i^{*^1}(s')+V_i^{*^2}(s'))}{\sum_{{s}'\in \mathcal{S}}V_i^{*^1}({s}')}\right.\\
			& \left. \quad -\sum_{s'\in \mathcal{S}}(V_i^{*^2}(s'))^2\frac{\sum_{s'\in \mathcal{S}}V_i^{*^1}(s')-V_i^{*^2}(s')}{\sum_{{s}'}V_i^{*^1}({s}')\sum_{{s}'\in \mathcal{S}}V_i^{*^2}({s}')} \right |\\
			&\leq \max _{s, a_i}\gamma \left |\sum_{s'\in \mathcal{S}}(V_i^{*^1}(s')-V_i^{*^2}(s'))\right|\cdot\frac{1}{\sum_{{s}'\in \mathcal{S}}V_i^{*^1}({s}')} \\
			& \quad \cdot\max \left| (V_i^{*^1}(s')+V_i^{*^2}(s'))-\frac{\sum_{s'\in \mathcal{S}}(V_i^{*^2}(s'))^2}{\sum_{{s}'\in \mathcal{S}}V_i^{*^2}({s}')} \right |\\
			&\leq \gamma |\mathcal{S}| \left\| Q_{i}^1-Q_{i}^2\right\|_{\infty}\cdot\frac{1}{|\mathcal{S}|\eta r_{\min}}\cdot\eta(2r_{\max}-r_{\min})\\
			&= \gamma (\frac{2r_{\max}}{r_{\min}}-1) \left\| Q_{i}^1-Q_{i}^2\right\|_{\infty}.
			\end{align*}
		\end{small}
		The third term of the penultimate line is because: if $V_i^{*^1}(s')+V_i^{*^2}(s') > \frac{\sum_{s'\in \mathcal{S}}(V_i^{*^2}(s'))^2}{\sum_{{s}'\in \mathcal{S} }V_i^{*^2}({s}')}$,
		\begin{small}
			\begin{align*}
			&V_i^{*^1}(s')+V_i^{*^2}(s') - \frac{\sum_{s'\in \mathcal{S}}(V_i^{*^2}(s'))^2}{\sum_{{s}'\in \mathcal{S}}V_i^{*^2}({s}')} \\
			&\leq V_i^{*^1}(s')+V_i^{*^2}(s') - \frac{\sum_{s'\in \mathcal{S}}(V_i^{*^2}(s'))*\eta r_{\min}}{\sum_{{s}'\in \mathcal{S}}V_i^{*^2}({s}')}
			\leq 2\eta r_{\max} - \eta r_{\min},
			\end{align*}
		\end{small}
		else,
		\begin{small}
			\begin{align*}
			&\frac{\sum_{s'\in \mathcal{S}}(V_i^{*^2}(s'))^2}{\sum_{{s}'\in \mathcal{S}}V_i^{*^2}({s}')} - (V_i^{*^1}(s')+V_i^{*^2}(s')) \\
			& \quad \leq \frac{\sum_{s'\in \mathcal{S}}(V_i^{*^2}(s'))*\eta r_{\max}}{\sum_{{s}'\in \mathcal{S}}V_i^{*^2}({s}')} \leq \eta r_{\max}.
			\end{align*}
		\end{small}
		Since $2\eta r_{\max} - \eta r_{\min} \geq \eta r_{\max}$, we have 
		\begin{small}
			$$| (V_i^{*^1}(s')+V_i^{*^2}(s'))-\frac{\sum_{s'\in \mathcal{S}}(V_i^{*^2}(s'))^2}{\sum_{{s}'\in \mathcal{S}}V_i^{*^2}({s}')}| \leq 2\eta r_{\max} - \eta r_{\min}.$$ 
		\end{small}
		Therefore, if $\gamma < \frac{r_{\min}}{2r_{\max}-r_{\min}}$, the operator $\mathcal{T}$ is a contraction. By contraction mapping theorem, $\mathcal{T}$ converges to a unique fixed point. 
	\end{proof}
\end{Theorem}


\section{Settings and Hyperparameters} \label{appendix:hyperparameters}

\begin{figure}
	\centering
	\begin{subfigure}{.2\linewidth}
		\includegraphics[width=1\linewidth]{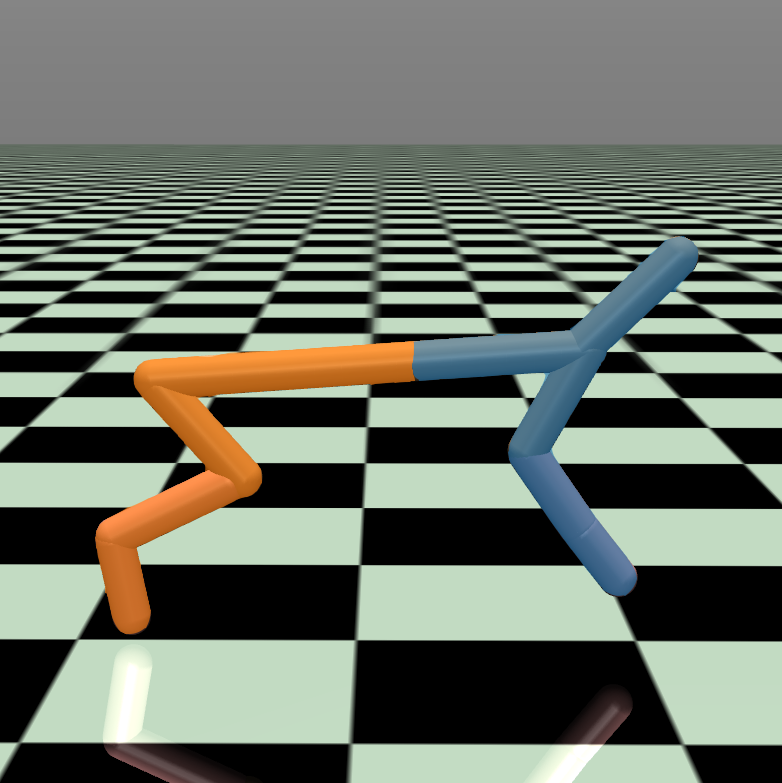}
	\end{subfigure}
	\begin{subfigure}{.2\linewidth}
		\includegraphics[width=1\linewidth]{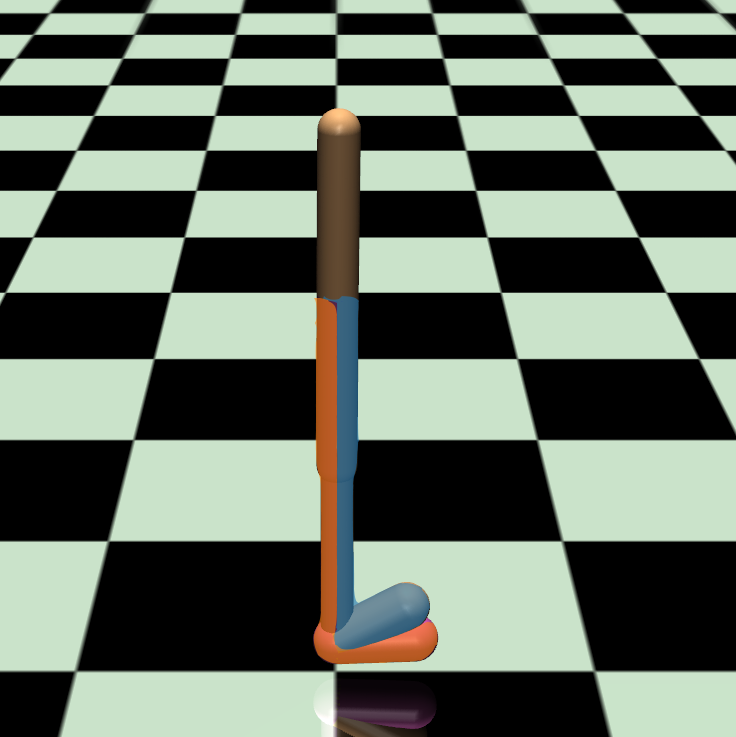}
	\end{subfigure}
	\begin{subfigure}{.2\linewidth}
		\includegraphics[width=1\linewidth]{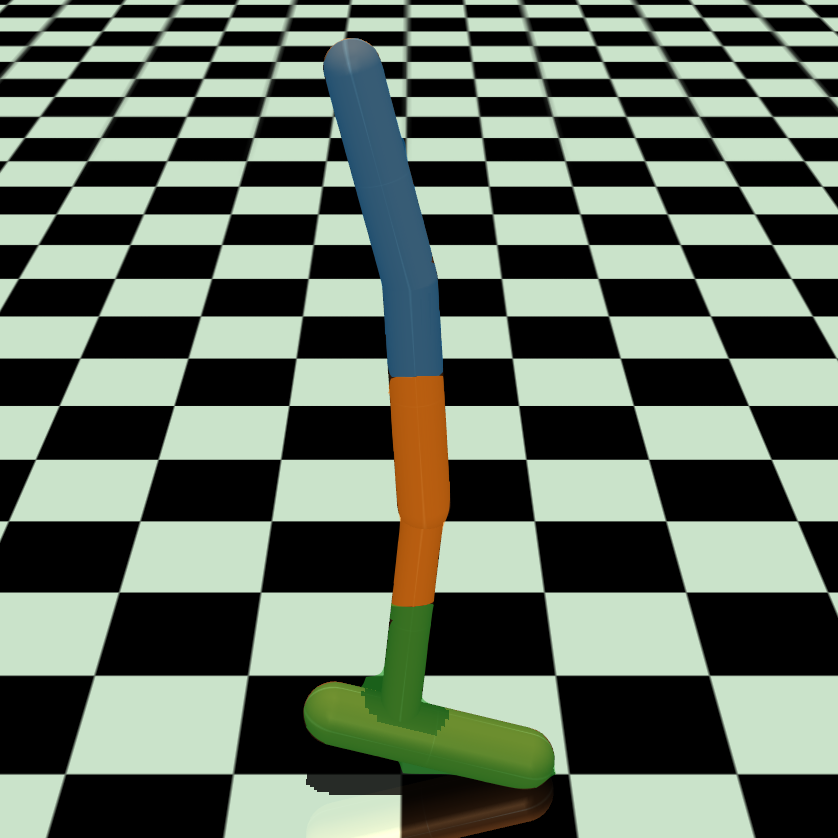}
	\end{subfigure}
	\begin{subfigure}{.2\linewidth}
		\includegraphics[width=1\linewidth]{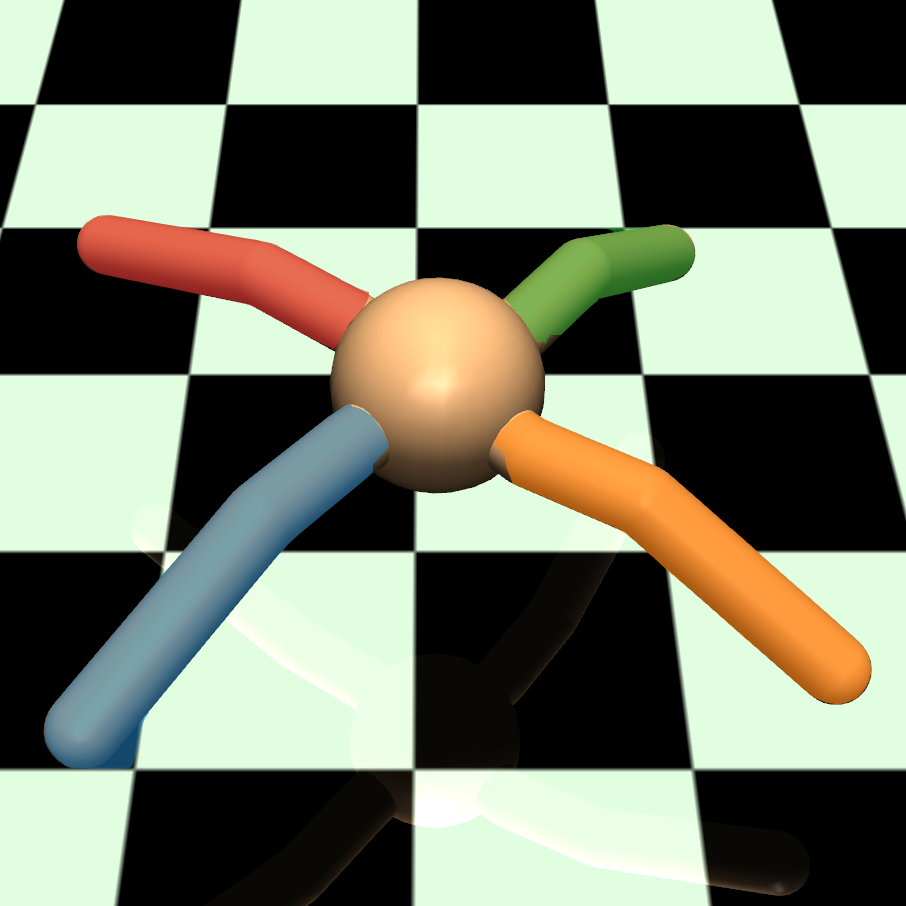}
	\end{subfigure}
	\caption{Illustrations of multi-agent Mujoco: HalfCheetah, Walker, Hooper, and Ant.}
	\label{fig:scenarios}
\end{figure}

\begin{table*}[t]
	\renewcommand{\arraystretch}{1}
	\centering
	\caption{Hyperparameters}
	\label{tab:hyperparameter}
	\begin{small}
		\begin{tabular}{@{}ccccc@{}}
			\toprule
			Hyperparameter&continues BCQ&discrete BCQ&CQL&TD3+BC\\
			\midrule
			
			learning rate of $Q$ &{$10^{-3}$} & $10^{-4}$& $10^{-4}$& $3 \times 10^{-4}$\\
			learning rate of $\xi$ &{$10^{-4}$} & & &\\
			learning rate of $G$ &{$10^{-4}$} & & &\\
			learning rate of $\pi$ & & & {$10^{-4}$}&$3 \times  10^{-4}$\\
			$\Phi$ &{$0.05$} & &\\
			$n$  &{$10$} & &\\
			VAE hidden space   &{$10$}& &\\
			$\alpha$ & & &$0.2$ & \\
			$\lambda$ & & & & $2.5$\\
			threshold  & & $\frac{0.6}{\text{action space}}$ & & \\
			\bottomrule
		\end{tabular}
	\end{small}
\end{table*}

The illustrations of multi-agent mujoco are shown in Figure~\ref{fig:scenarios}, different colors indicate different agents. Each agent independently controls one or some joints of the robot and can get the state and reward of the robot, which are defined in the original tasks. For each environment, we collect $N$ datasets for the $N$ agents. Each dataset contains $1\times 10^6$ transitions $(s,a_i,r,s',done)$. For data collection, we train an intermediate policy and an expert policy for each agent using SAC \cite{haarnoja2018soft}. The offline dataset $\mathcal{B}_i$ is a mixture of four parts: $20\%$ transitions are split from the experiences generated by the SAC agent at the early training, $35\%$ transitions are generated from that the agent $i$ acts the intermediate policy while other agents act the expert policies, $35\%$ transitions are generated from that agent $i$ performs the expert policy while other agents act the intermediate policies, $10\%$ transitions are generated from that all agents perform the expert policies. For the last three parts, we add a small noise to the policies to increase the diversity of the dataset.

In all tasks, we set the discount factor $\gamma = 0.99$ and use ReLU activation. In Mujoco tasks, the MLP units are $(64,64)$, and the batch size is $1024$. In SMAC and DG, the MLP units are $(256,256)$, and the batch size is $100$. The hyperparameter of our framework is the optimism level $\epsilon$. We respectively set $\epsilon = 0.80, 0.48,0.80,0.64$ in HalfCheetah, Walker, Hopper, and Ant, set $\epsilon = 0.99$ in SMAC, and set $\epsilon = 0.9$ in MPE. The hyperparameters of baselines are summarized in Table~\ref{tab:hyperparameter}. 

In this paper, we use SMAC (MIT license), MPE (MIT license), and Gym (MIT license), D4RL (Apache-2.0 license). Many thanks for their contributions.

\end{document}